\pdfoutput=1

\documentclass[11pt]{article}

\usepackage[]{EMNLP2023}

\usepackage{marvosym}
\usepackage{times}
\usepackage{enumerate}
\usepackage{latexsym}
\usepackage{todonotes}
\usepackage{float}
\usepackage{tikz}
\usepackage{graphicx}
\usepackage{multirow}
\usepackage{marvosym}
\usepackage{amssymb}
\usepackage{pifont}
\newcommand{\cmark}{\ding{51}}%
\newcommand{\xmark}{\ding{55}}%
\usepackage{comment}
\usepackage{hyperref}

\usepackage[T1]{fontenc}

\usepackage[utf8]{inputenc}

\usepackage{microtype}

\usepackage{inconsolata}
\usepackage{amssymb}
\usepackage{amsmath}
\usepackage{amsthm}
\usepackage{algorithm}
\usepackage{algpseudocode}
\usepackage{multirow}
\usepackage{booktabs} 
\usepackage{float}
\usepackage{natbib}
\usepackage{arydshln}
\usepackage{soul}

\newtheorem{definition}{Definition}

\newcommand{\docset}{\ensuremath{\math{\textbf{D}}}}
\newcommand{\doc}{\ensuremath{d}}
\newcommand{\summset}{\ensuremath{\math{\textbf{S}}}}
\newcommand{\summ}[1]{\ensuremath{\math{s_{#1}}}}

\newcommand{\profileset}{\ensuremath{\math{\textbf{U}}}}
\newcommand{\profile}{\ensuremath{u}}
\newcommand{\model}{\ensuremath{\math{M_{\boldsymbol{\theta},u}}}}

\newcommand{\egises}{{\texttt{EGISES}}}
\newcommand{\copernicus}{{\texttt{iCOPERNICUS}}}

\newcommand{\contrastiveprompt}{\ensuremath{\math{\mathcal{P_C}}}}  

\newcommand{\contrastivedemo}{\ensuremath{\math{\mathcal{\mathcal{D}_{\mathcal{P_C}}}}}}

\newcommand{\contrastivedemoICPLexample}[3]{\ensuremath{\math{(\bigoplus\limits_{j=1}^{#3}h_{#2})\oplus (\bigoplus\limits_{i=1}^{n} (d_{#1} \oplus \bigoplus\limits_{j=1}^{#3} u_{#1#2}))}}}

\newcommand{\contrastivedemoinputlabel}[3]{\ensuremath{\math{ s_{(\doc_{#1},h_{#2},h_{#3})}^* \equiv \langle u_{#1#2}\oplus u_{#1#3}\rangle}}} 

\newcommand{\contrastivedemoquery}[3]{\ensuremath{\math{\doc_q}}} 

\newcommand{\degress}{{\texttt{DEGRESS}}}

\newcommand{\summleveldegress}[2]{\ensuremath{\math{\degress(s_{u_{#1#2}}|(d_{#1}, u_{#1#2}))}}}

\newcommand{\usercondweightequation}[3]{\ensuremath{\math{{w(u_{#1#2} | u_{#1#3})  = \frac{\sigma(u_{#1#2} , u_{#1#3})}{\sigma(u_{#1#2} , d_{#1})}}}}}
\newcommand{\summcondweightequation}[3]{\ensuremath{\math{{w({s_{u_{#1#2}}} | s_{u_{#1#3}})  = \frac{\sigma(s_{u_{#1#2}} , s_{u_{#1#3}})}{\sigma(s_{u_{#1#2}} , d_{#1})}}}}}

\newcommand{\userconddevequation}[4]{\ensuremath{\math{\frac{\exp(w(u_{#1#2} | u_{#1#3}))}{\sum\limits_{#4=1}^{|\textbf{U}_{d_{#1}}|}{\exp(w(u_{#1#2} | u_{#1#4}))}} \cdot \sigma(u_{#1#2} , u_{#1#3})}}}
\newcommand{\summconddevequation}[4]{\ensuremath{\math{\frac{\exp(w(s_{u_{#1#2}} | s_{u_{#1#3}}))}{\sum\limits_{#4=1}^{|\textbf{U}_{\doc_{#1}}|}{\exp(w(s_{u_{#1#2}} | s_{u_{#1#4}}))}} \cdot \sigma(s_{u_{#1#2}} , s_{u_{#1#3}})}}}

\title{In-Context Personalization Learning in LLMs for Text Summarization Task}
\title{\copernicus: Evaluating In-Context Personalization Learning in LLMs for Text Summarization Task}
\title{Summarizer LLMs may not exhibit In-Context Personalization Learning}
\title{Do LLMs exhibit In-Context Personalization Learning? Maybe not!}
\title{Are LLMs smart enough for In-Context Personalization Learning?}
\title{\textit{Are Large Language Models In-Context Personalized Summarizers?} \\ Get an \copernicus\ Test Done!}

\author
{
    Divya Patel $^{\dagger *} \quad$ Pathik Patel $^{\dagger *} \quad$ Ankush Chander ${ }^{\dagger *}\quad$ Sourish Dasgupta ${ }^{\dagger *} \quad$ Tanmoy Chakraborty $^{\ddagger}$\\
    ${ }^{\dagger}$ KDM Lab, Dhirubhai Ambani Institute of Information \& Communication Technology, India\\
    $\ddagger$ Indian Institute of Technology, Delhi, India\\
    \{202001420, 202003002, ankush\_chander, \Letter \ sourish\_dasgupta\}@daiict.ac.in, \Letter \ tanchak@iitd.ac.in
}

\let\svthefootnote\thefootnote
\newcommand\blfootnotetext[1]{%
  \let\thefootnote\relax\footnote{#1}%
  \addtocounter{footnote}{-1}%
  \let\thefootnote\svthefootnote%
}

\let\svfootnotetext\footnotetext
\renewcommand\footnotetext[2][?]{%
  \if\relax#1\relax%
    \ifnum\value{footnote}=0\blfootnotetext{#2}\else\svfootnotetext{#2}\fi%
  \else%
    \if?#1\ifnum\value{footnote}=0\blfootnotetext{#2}\else\svfootnotetext{#2}\fi%
    \else\svfootnotetext[#1]{#2}\fi%
  \fi
}

\begin{document}
\maketitle
\footnotetext{*Equal contributions.}
\begin{abstract}
Large Language Models (LLMs) have succeeded considerably in In-Context-Learning (ICL) based summarization. However, saliency is subject to the users' specific preference histories. Hence, we need reliable {\em In-Context \textbf{\underline{Personalization}} Learning} (IC\textbf{P}L) capabilities within such LLMs. For any arbitrary LLM to exhibit ICPL, it needs to have the \textbf{ability to discern contrast in user profiles}. A recent study proposed a measure for \textit{degree-of-personalization} called EGISES for the first time. EGISES measures a model's responsiveness to user profile differences. However, it cannot test if a model utilizes all three types of cues provided in ICPL prompts: (i) example summaries, (ii) user's reading histories, and (iii) contrast in user profiles. To address this, we propose the \copernicus\ framework, a novel \textbf{I}n-\textbf{Co}ntext \textbf{Per}sonalization Lear\textbf{ni}ng S\textbf{c}r\textbf{u}tiny of \textbf{S}ummarization capability in LLMs that uses EGISES as a comparative measure. As a case-study, we evaluate 17 state-of-the-art LLMs based on their reported ICL performances and observe that 15 models' ICPL degrades (min: 1.6\%$\downarrow$; max: 3.6\%$\downarrow$) when probed with richer prompts, thereby showing lack of \textit{true} ICPL.
\end{abstract}

\section{Introduction}\label{sec: introduction}
With the constant influx of information, we need efficient models capable of summarizing essential content from lengthy documents for faster comprehension and prioritization \cite{personalized-summarization-utility}. Yet, defining what constitutes ``salient'' information remains subjective, particularly in documents covering multiple aspects. To tackle this, contemporary summarizers should be personalized to users' preference histories and interests.

\textbf{Specialized PLMs as summarizers.} \citet{pens} proposed the most direct method to train models to learn personalization using user reading histories. These models (called PENS models) use variants of pointer-generator networks \cite{pointer-gen} that are injected with representations of user reading history for user preference alignment. Other indirect approaches include aspect-based models \citep{aspect-based-4,aspect-based-2} that produce summaries coherent with the document themes but lack adaptability to changes in the reader's profile. In contrast, interactive human-feedback-based models allow for iterative refinement based on user feedback, thereby better personalization \cite{adaptivesum}. 

\textbf{LLMs as personalized summarizers.} 
Recent studies on the state-of-the-art (SOTA) LLMs show unprecedented In-Context Learning (ICL) based summarization performance \citep{ICL-summarization-news,ICL-summarization-meeting,ICL-summarization-dialogue}. This opens the possibility of In-Context Personalization Learning (ICPL) in these LLMs. At the same time, it also underscores the necessity for robust and dependable methods of evaluating the degree of ICPL within such models. Although benchmarking of LLMs for summarization has been done for accuracy, fluency, and consistency \citep{LLM_Sum_Eval}, so far, no study has been done yet on the probing and evaluation of ICPL in LLMs for the summarization task. In this paper, we propose \copernicus, an \textbf{I}n-\textbf{Co}ntext \textbf{Per}sonalization Lear\textbf{ni}ng S\textbf{c}r\textbf{u}tiny of \textbf{S}ummarization capability in LLMs. 

\textbf{\copernicus\ framework.} \copernicus\ is a prompt-based probing framework that investigates whether LLMs exhibit true ICPL using a 3-pronged approach: (i) whether few-shot prompting of \textbf{examples (i.e., reader-generated gold references) enhances ICPL}, (ii) whether adding reader's \textbf{reading history improves ICPL}, and (iii) whether \textbf{contrastive profile information} showing subjective differences in reader-preferences for the same document \textbf{induce better ICPL}. Since \copernicus\ is a \textit{comparative} framework, it needs a personalization measure for analyzing the influence of the injected profile information in the prompts. We use EGISES-JSD, the only known measure for personalized summarization \cite{egises}. EGISES measures the ability of models to discern the differences in user profiles and generate summaries that are proportionately different. 

\begin{table}[t] \normalsize
\centering
    \renewcommand*{\arraystretch}{1}
    \scalebox{0.7}
    {\begin{tabular}{llccc}
    \hline
        \multicolumn{5}{c}{\textbf{Model Variants Probed}}\\
        \hline
        \multirow{1}{*}{\textbf{Base-}} & \multirow{1}{*}{Llama 2 (7B, 13B) \citep{llama-2-base-underfitting}} \\
        \multirow{1}{*}{\textbf{models}} &\multirow{1}{*}{Mistral v0.1 (7B) \citep{mistral-(GQA-SWA)}}\\ 
        \hline
        \multirow{3}{*}{\textbf{Instruct-}} &\multirow{1}{*}{Mistral 7B Instruct v0.1 \citep{mistral-(GQA-SWA)}} \\
        &\multirow{1}{*}{Mistral 7B Instruct v0.2 \citep{mistral-(GQA-SWA)}}\\

        \multirow{1}{*}{\textbf{tuned}}&\multirow{1}{*}{Tulu V2 (7B, 13B) \citep{Tulu}} \\
        
        &\multirow{1}{*}{Orca 2 (7B, 13B) \citep{orca-2}}\\
        
        &\multirow{1}{*}{Stable Beluga (7B, 13B) \citep{stable-beluga-2}}\\
        
        \hline
         
        \multirow{1}{*}{\textbf{RLHF-tuned}} & \multirow{1}{*}{Llama 2 Chat (7B, 13B) \citep{llama-2-base-underfitting}}\\
        \hline
        
        \multirow{3}{*}{\textbf{DPO-tuned}} &\multirow{1}{*}{Tulu V2 DPO (7B, 13B) \citep{Tulu}}\\
        
        &\multirow{1}{*}{Zephyr 7B $\alpha$ \citep{Zephyr}}\\
        &\multirow{1}{*}{Zephyr 7B $\beta$ \citep{Zephyr}} \\
        
        \hline
    \end{tabular}}
    \caption{LLMs probed for ICPL w.r.t summarization.}
    \label{tab:LLMs Probed}
    \vspace{-5mm}
\end{table}

\textbf{Observations} As a case-study of the application of \copernicus, we probe ten SOTA LLMs that exhibit reasonably good ICL on standard benchmark tasks, six of which have 7B and 13B model size variants (see Table \ref{tab:LLMs Probed}). We use the PENS dataset \citep{pens} as in \citep{egises} to compare the ICPL results with the baseline specialized personalized summarization models evaluated therein. We observe that all the studied models, except for Orca-2 7B and Zephyr 7B $\beta$, exhibit performance degradation in all the three probes of \copernicus\ - i.e., injection of examples, history, and contrastive profile information. 

\textbf{Key Contributions.} We present for the first time: (i) a detailed introduction of ICPL for personalized summarization, (ii) \copernicus\ as a formal framework of evaluation of ICPL in LLMs, and (iii) a detailed case-study of the application of \copernicus\ tests for determining ICPL in the selected SOTA LLMs.

\section{Preliminaries}\label{sec: prelims}
\subsection{Personalization w.r.t Summarization} \label{sec: personalization}
As proposed in \citet{egises}, the \textit{degree-of-personalization} is a quantitative measure of how much a summarization model fine-tuned for personalization is adaptive to a user's (i.e., reader's) subjective expectation. This also implies that it measures how accurately a model can capture the \textit{ user's "evolving" \textbf{profile} reflected through \textbf{reading history}} (i.e., a temporal span of the reading and skipping actions of a user on a sequence of documents that is interleaved by the actions of generating and reading summaries). This is because the \textit{subjective expectation is a function of the reading history}. \textbf{A low degree of personalization, by definition, implies poor user experience (UX)}. If a model does not efficiently capture the user's profile, it may lead to irrelevant summaries. In this situation, poor UX would mean that the user would have to spend more time getting to the information he/she is interested in or suffer from information overload and fatigue. However, this irrelevant information can be useful for a different user with a different profile. To illustrate this, we borrow the example given by \citet{egises} where if reader Alice, who has been following "\textit{civilian distress}" in the Hamas-Israeli conflict, reads a news summary whose content is primarily about "\textit{war-front events}", her UX will drop down due to information overload and high time-to-consume, even though her interest is also covered to a fair extent. In contrary, a reader Bob, who has been mostly following war news, would have quite high UX.

\subsection{EGISES: Personalization Measure} \label{sec: personalization vs. accuracy}
\citet{egises} showed theoretically and empirically that a model could have high accuracy scores in both the examples in the previous section, although the individual degree of personalization differs. This can \textit{mislead selection of a model for a fairly high accuracy score even though it suffers from poor UX}. To address this, they proposed EGISES as, to the best of our knowledge, the only known measure for personalized summarization evaluation. Informally, EGISES measures the extent to which a model can discern the differences between user profiles and generate summaries that have proportionate differences. Difference is modeled as a chosen divergence on a metric space. See Appendix \ref{app: egises} for formulation.

\subsection{In-Context Personalization Learning}\label{sec: ICPL for summarization}

In-Context learning (ICL) is an emergent phenomenon exhibited LLMs (first highlighted in \citet{ICL-few-shot-GPT3} for GPT-3), where models acquire proficiency in unknown tasks on which they are not pre-trained from limited examples, called {\em prompts}, with no update in their parameters (i.e., the models are frozen).\footnote{For formalizations of ICL see Appendix \ref{app: ICL description}.} In-Context Personalization Learning (ICPL) for summarization is a special case of ICL where, for a document query $d_q$, an LLM is expected to generate the user-specific optimal summary $s^*_{(d_q, h_j)}$, for the $j$-th user expecting the summary of $d_q$. Here, $h_j$ is the user's \textbf{reading history} (temporal sequence of the user's click and skip history of documents). $s^*_{(d_q, h_j)}$ is the same as the $j$-the user's expected summary $u_{qj}$ (i.e., gold-reference), and hence can be denoted $s^{*}_{u_{qj}}$.


\begin{definition} \label{def:prompt4ICPL}
    \textbf{Prompt4ICPL}: A prompt $\mathcal{P}_{ICPL}$ to a language model $M$ consists of an user's reading history ($h$), an optional sequence of $n$ concatenated ($\oplus$) demonstration examples (i.e., input-label pairs) as: $h_{j}\oplus \bigoplus\limits_{i=1}^{n} (d_{i} \oplus u_{ij})$ where, $d_i$ is the example document to be summarized for $j$-th user ($u_{ij}$ being the \textbf{gold-reference summary example}), and a query document $d_q$, such that $d_{i} \neq d_q$.
\end{definition}

A \textbf{zero-shot prompt (0-shot)} is the special case when demonstration examples are not provided (i.e., $u_{ij}=\emptyset$) while $d_q$ and the user reading history $h_j$ are given in the prompt. The few-shot version can be of two types: (i) \textbf{with history (k-shot w/hist.)}, and (ii) \textbf{without history (k-shot w/o hist.)}. In the second type, the user profile is not represented by reading history but rather by the expected summaries (or gold-reference summaries). This can seen as the user's ``\textit{writing history}". 

\section{The \copernicus\ Framework}\label{sec:copernicus-framework}
\copernicus\ is a novel prompt-based three-pronged probing framework for evaluating ICPL in LLMs. It tests \textit{whether the model can harness \textbf{three types of profile information} included within the test prompts: (i) \textbf{examples}, (ii) \textbf{history}, and (iii) \textbf{contrastive information} (in terms of history and examples)}. Before we provide a detailed outline of the framework, we first discuss the contrastive probing setup in the following section.

\subsection{Contrastive Probing}\label{sec:degree-of-ICPL}
One of the key probes of the \copernicus\ framework is testing LLMs for ICPL with \textit{contrastive examples}, i.e., input-label pairs containing at least two user (i.e., reader) profiles (can be reading or writing history) with the query document $d_q$. An LLM capable of ICPL should be able to discern the difference between the reader profiles and generate summaries accordingly (i.e., $s_{u_{ij}}$ and $s_{u_{ik}}$) in line with the notion of insensitivity-to-subjectivity as defined by \citet{egises} (see Appendix \ref{app:insensitivity}). Contrastive Prompt4ICPL ($\mathcal{P_C}$) is defined as:   

\begin{figure}[t]
\centerline{\includegraphics[width= 6.9cm,scale=1]{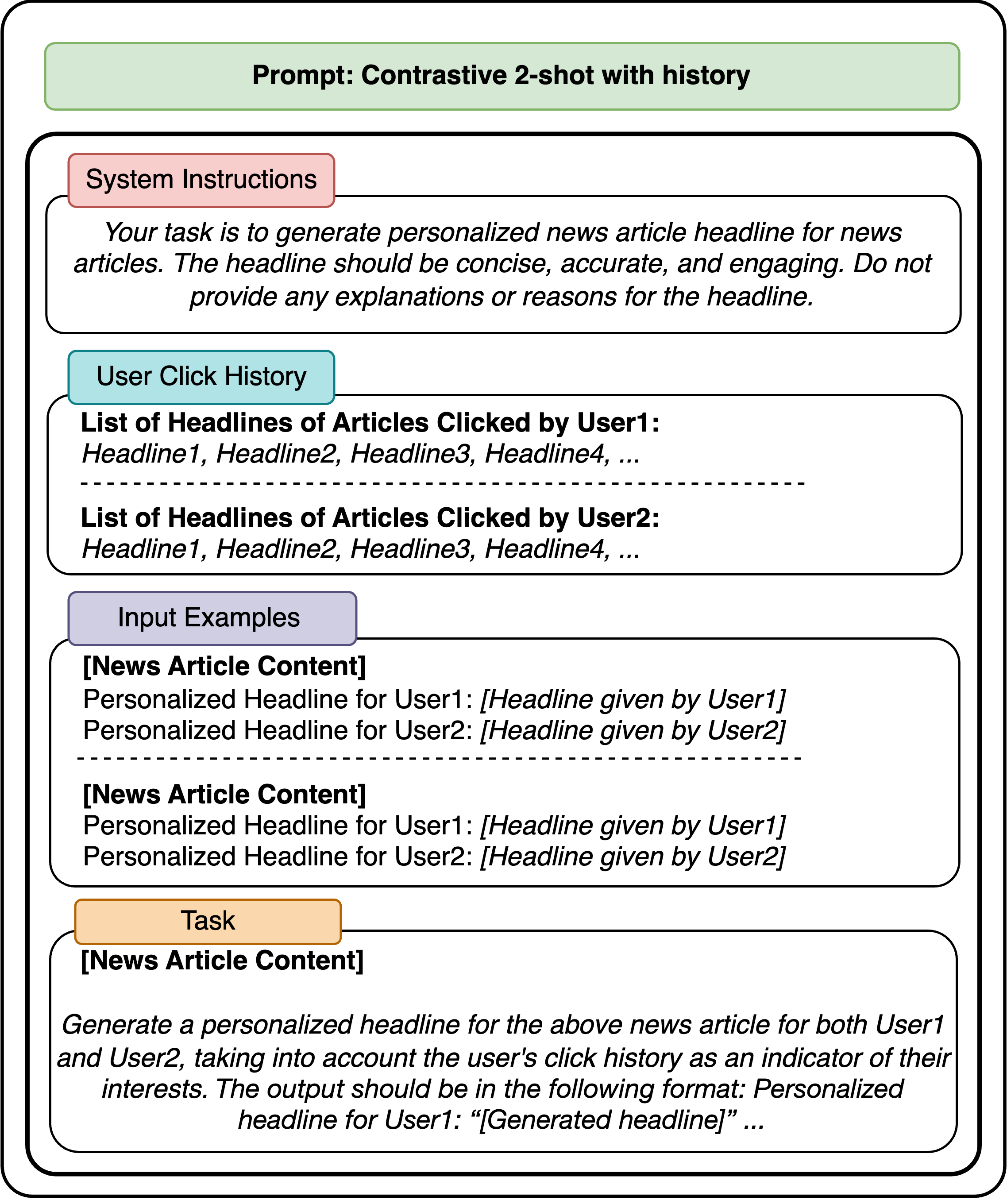}}
\vspace{-1.5mm}
\caption{$\mathcal{P_C}$ type: Contrastive (C)-2-shot w/ history.}
\label{fig:C-2-shot-w/ hist.}
\vspace{-5mm}
\end{figure}

\begin{definition} \label{def:contrastive-prompt}
    \textbf{Contrastive Prompt4ICPL ($\mathcal{P_C}$)}: A $\mathcal{P_C}$ given to a language model $M$ is a sequence of $n$ concatenated ($\oplus$) \textbf{contrastive} demonstration examples $\mathcal{D}_{\mathcal{P_C}}$ as: $(\bigoplus\limits_{j=1}^{m}h_{j})\oplus (\bigoplus\limits_{i=1}^{n} (d_{i} \oplus \bigoplus\limits_{j=1}^{m} u_{ij}))$), each having $m$ \textbf{subjective} expected summaries ($u_{ij}$), and a query document $d_q$, s.t. $d_{i} \neq d_q$. 
\end{definition}

A \textbf{contrastive zero-shot prompt (C-0-shot)}, contains $h_{j}$ and $h_k$ representing the contrastive reading-histories of two users $j$ and $k$ with no demonstration examples. The few-shot version, similar to k-shot (plain) prompt, is of two types: (i) \textbf{C-k-shot w/hist.} (Figure \ref{fig:C-2-shot-w/ hist.}) and (ii) \textbf{C-k-shot w/o hist}. In the second case, gold-reference summaries (writing history) of \textbf{both users} are given in the prompt as examples but not their reading histories.\footnote{NT: {\textit{For sake of clarity, contrastive prompts are not chain-of-thought (CoT) prompts as it does not contain thought break-down or require thought generation}}.} We now define ICPL (weak and strong cases) as:

\begin{definition} \label{def:weak ICPL}
    \textbf{Weak ICPL.} Given a contrastive prompt $\mathcal{P_C}$, an LLM $M_{\boldsymbol{\theta},u}$ exhibits weak ICPL, if $\forall(u_{qj}, u_{qk})$ w.r.t query document $d_q$, $(\sigma(u_{qj}, u_{qk})\leq \tau_{max}^{U}) \iff
(\sigma(s_{u_{qj}}, s_{u_{qj}}) \leq \tau_{max}^{S})$; $\tau_{max}^{U}, \tau_{max}^{S}$ are bounds within which users' expected (gold-reference) summary pair $(u_{qj}, u_{qk}$) and LLM-generated summary pair ($s_{u_{qj}}, s_{u_{qj}}$) are indistinguishable; $\sigma$ is an arbitrary distance metric on the metric space $\mathcal{M}$, where $d, u, s$ are defined. 
\end{definition}

\begin{definition} \label{def:strong ICPL}
    \textbf{Strong ICPL.} Given $\mathcal{P_C}$, $M_{\boldsymbol{\theta},u}$ exhibits strong ICPL, if $\forall(u_{qj}, u_{qk})$ w.r.t $d_q$, $M_{\boldsymbol{\theta},u}$ satisfies: (i) weak ICPL, and (ii) $(\sigma(u_{qj}, u_{qk}) > \tau_{max}^{U}) \iff (\sigma(s_{u_{qj}}, s_{u_{qj}}) > \tau_{max}^{S})$.
\end{definition}
The following sections describe the three-pronged \copernicus\ framework. 

\subsection{Probe 1: Do example summaries help?}
The first probe within the \copernicus \ framework studies the impact of k-shot prompts (in contrast to 0-shot prompts) on LLM models.\footnote{It is to be noted that, as pointed out in Section \ref{sec:degree-of-ICPL}, 0-shot prompts contain the reading histories ($h$).} The (plain) k-shot w/o hist. prompt-based probing investigates whether the model improves ICPL performance w.r.t EGISES-scores by mapping the key latent concepts in each example summary with those in the corresponding document for any given user. This is a much richer cue than the 0-shot case, where the model does not get much assistance but can only observe inter-document conceptual associations (i.e., user's click and skip patterns) provided as a part of the reading history $h$. A richer version of this probe is that with k-shot w/ hist. prompts, where additional history information is also provided to investigate if the model can associate that with the examples provided. These two probes should also be performed in the contrastive prompt setting (C-k-shot prompts (w/ and w/o hist.)) to investigate the presence of the same kind of associations with the additional capacity of associating user-specific profile concepts. More specifically, ideally, a model should be able to \textbf{harness example summaries} in the following order of ICPL performance:
\begin{enumerate}
    \item (plain) 0-shot $\prec$ \textbf{(plain) \underline{k-shot} w/o hist.}, violation leads to \textbf{Paradox 1 (PX-1)}.
    \item (plain) 0-shot $\prec$ \textbf{(plain) \underline{k-shot} w/ hist.}, violation leads to \textbf{Paradox 1 w/ hist. (PX-1-h)}.
    \item C-0-shot $\prec$ \textbf{C-\underline{k-shot} w/o hist.}, violation leads to \textbf{PX-1 (contrastive)}.
    \item C-0-shot $\prec$ \textbf{C-\underline{k-shot} w/ hist.}, violation leads to \textbf{PX-1-h (contrastive)}.
\end{enumerate}  

\subsection{Probe 2: Does reading-history help?}
The second probe investigates whether a model can utilize the temporal sequence of a specific user's reading (i.e., document clicking and skipping) history and associate the innate latent concepts with that of the corresponding example summaries provided in the prompt (both plain and contrastive). Hence, a model should be able to \textbf{harness reading history} in the following ICPL performance order:

\begin{enumerate}
    \item (plain) k-shot w/o hist. $\prec$ \textbf{(plain) k-shot \underline{w/ hist.}}, violation leads to \textbf{PX-2}.
    \item C-k-shot w/o hist. $\prec$ \textbf{C-k-shot \underline{w/ hist.}}, violation leads to \textbf{PX-2 (contrastive)}.
\end{enumerate}  

\subsection{Probe 3: Do contrastive prompts help?}
The third probe investigates whether models can capitalize on additional contrasting (i.e., similarity/differences) information about user profiles in contrastive prompts. Hence, a model should be able to \textbf{harness contrastive information} in the following order of ICPL performance:

\begin{enumerate}
    \item (plain) 0-shot $\prec$ \textbf{\underline{C}-0-shot}, violation leads to \textbf{PX-3}.
    \item (plain) k-shot w/o hist. $\prec$ \textbf{\underline{C}-k-shot w/o hist.}, violation leads to \textbf{PX-4}.
    \item (plain) k-shot w/ hist. $\prec$ \textbf{\underline{C}-k-shot w/ hist.}, violation leads to \textbf{PX-5}.
\end{enumerate}  



\begin{table*}[t]
    \centering
    \scalebox{0.8}
    {
    \begin{tabular}{lcccc}
        \toprule
        \hline
        \textbf{Prompt Style} & \textbf{Reading Hist.} & \textbf{Examples} & \textbf{Article Body} & \textbf{\# Prompts}\\
        \midrule
        \hline
        0-shot & 1200 Tokens & -- & 2500 Tokens & 6856\\
        C-0-shot & 1000 x 2 Tokens & -- & 1700 Tokens& 5246 \\
        2-shot w/o hist. & -- & 950 x 2 Tokens& 1800 Tokens& 6798\\
        C-2-shot w/o hist. & -- & 950 x 2 Tokens & 1800 Tokens & 5246\\
        2-shot w/ hist. & 1200 Tokens & 600 x 2 Tokens & 1300 Tokens & 6798\\
        C-2-shot w/ hist. & 850 x 2 Tokens & 450 x 2 Tokens & 1100 Tokens & 5246\\
        \hline
        \bottomrule
    \end{tabular}
    }
    \caption{\copernicus\ prompt composition (w.r.t \# of tokens) for all prompt styles; \textbf{NT}: overall prompt size is fixed.}
    \vspace{-5mm}
    \label{tab:ContextLengthTable}
\end{table*}

\subsection{Limitations of EGISES w.r.t ICPL}\label{sec: egises limitation}
As can be seen from the previous section, \copernicus\ is a \textit{comparative} framework that needs a personalization measure for analysis of the influence of profile information. Since it was reported in \citet{egises} that the EGISES-JSD variant performed well regarding human-judgment correlation, we choose this specific variant as the comparative measure within the \copernicus\ framework to evaluate system-level strong degree-of-ICPL (i.e., $\sigma$ in definitions \ref{def:weak ICPL} and \ref{def:strong ICPL} is JSD (Jensen-Shannon Divergence)).\footnote{NT: EGISES as a standalone measure \textbf{has not been designed to test ICPL} (see Appendix \ref{app: egises} for formulation).} However, it is to be noted that \copernicus\ is not tightly coupled with EGISES; any better future personalization evaluation measure can also be applied. In section \ref{sec: EGISES misleading}, we empirically show that absolute EGISES score-based ICPL leaderboards can be misleading.   

\section{Evaluation: Setup}

\subsection{Model Benchmarking Dataset}\label{sec: pens dataset}
Evaluating the selected models w.r.t \copernicus\ requires the test dataset to contain (i) example (and expected) gold summaries, (ii) user's reading history, and (iii) contrastive examples (i.e., \textit{subjective} gold summaries). We selected the test data sourced from the PENS dataset \citep{pens}\footnote{We comply with the Microsoft Research License Terms.} for our purpose since, to the best of our knowledge, it is the only dataset containing all the above three. 103 college students, having English as their native language, were invited as voluntary participants. A two-phase process was adopted to construct the test set. Initially, the participants selected at least 50 articles of personal interest from a pool of 1000 news articles, which were then sorted based on exposure time. \textbf{This formed their reading histories}. Subsequently, participants in the second phase created preferred headlines (gold references) for 200 news articles without prior knowledge of the original headlines. \textbf{This formed the set of examples and expected (personalized) summaries}. The two-stage process ensures an average of four gold-reference summaries per article, \textbf{thereby enabling contrastive prompts to be sampled out}. For details, see Appendix \ref{app: PENS Dataset details}.

\subsection{Probing Dataset Creation}
We engineer six distinct prompt templates in accordance with the \copernicus\ framework (see Table \ref{tab:ContextLengthTable}). The prompts were sampled from the PENS dataset (section \ref{sec: pens dataset}) with sample size of 3840 news articles such that the total number of tokens for all the settings were 3700 - i.e., \textbf{the overall prompt size remained constant}. This was done so that the probes were comparable in a controlled environment. Depending on the specific test, each prompt has been broken up into history, examples, and article body.\footnote{See Figure \ref{fig:PromptStructure} for the structure of the prompts and Figures \ref{fig:Example PX-1}-\ref{fig:Ex_PX5} in the Appendix for examples.} 
The dataset is released for research purposes at \href{https://drive.google.com/file/d/1pgyZJywA51cs641DPMoZB6ab7obyOLtT/view?usp=drive_link}{KDM-Lab\_iCOPERNICUS\_prompt-dataset\_v1.0}.


\subsection{Probed SOTA LLMs}
We probe ten SOTA LLMs with their 7B and 13B variants (see Tables \ref{tab:LLMs Probed} and \ref{tab:Copernicus evaluation}), totaling seventeen variants. Models are chosen based on their recency, training data diversity, and performance on key benchmark tasks requiring comparative reasoning.\footnote{Tasks: commonsense reasoning (e.g., Winogrande \citep{winogrande}, Hellaswag \citep{hellaswag}), math (e.g., GSM8k \citep{gsm8k}), code (e.g., MBPP \citep{mbpp-code}), and multi-task benchmarks (e.g., MMLU \citep{mmmlu-multi-task}, AGI Eval \citep{agi})} We could not evaluate 13B+ models due to compute resource constraints. However, the core contribution of the paper is the \copernicus\ framework itself which is \textit{applicable for selection decision of \underline{any sized model}}. Hence, the evaluations is primarily a \textit{\underline{case study} of the application of \copernicus}. Appendix \ref{app:LLMdesc} has model descriptions.

\begin{table*}[t]\normalsize
\centering
\scalebox{0.66}{
\begin{tabular}{llcccccc}
\hline
 & \multicolumn{1}{l}{\textbf{LLM Model Variants}} & \multicolumn{1}{c}{\textbf{0-shot}} & \multicolumn{1}{c}{\textbf{2-shot w/o hist.}} & \multicolumn{1}{c}{\textbf{2-shot w/ hist.}} & \multicolumn{1}{c}{\textbf{C-0-shot}} & \multicolumn{1}{c}{\textbf{C-2-shot w/o hist.}} & \multicolumn{1}{c}{\textbf{C-2-shot w hist.}}\\ \hline
        \multirow{3}{*}{\textbf{Base models}} & \multicolumn{1}{l}{Llama 2 7B} & \multicolumn{1}{c}{0.408} & \multicolumn{1}{c}{0.367} &  \multicolumn{1}{c}{0.367} & \multicolumn{1}{c}{0.408} &	\multicolumn{1}{c}{0.46} & \multicolumn{1}{c}{0.458}\\
        
        &\multicolumn{1}{l}{Llama 2 13B} & \multicolumn{1}{c}{0.412} & \multicolumn{1}{c}{	0.371} &  \multicolumn{1}{c}{0.357} &	0.418 &	0.5 & 0.474\\ 
         &\multicolumn{1}{l}{Mistral 7B v0.1} & \multicolumn{1}{c}{0.398} & \multicolumn{1}{c}{0.353} &  \multicolumn{1}{c}{0.354} & 	0.464 &	0.406	& 0.469\\
        \hline
         \multirow{8}{*}{\textbf{Instruction-tuned}} &\multicolumn{1}{l}{Mistral 7B Instruct v0.1} & \multicolumn{1}{c}{0.4} &  \multicolumn{1}{c}{	0.366} & 	0.378 &	0.395	& 0.405	& 0.399\\
        &\multicolumn{1}{l}{Mistral 7B Instruct v0.2} & \multicolumn{1}{c}{0.391} &  \multicolumn{1}{c}{0.348} & 	\textbf{0.354}	& 0.359 &	\textbf{0.339} &	0.369\\

        
        &\multicolumn{1}{l}{Tulu V2 7B} & \multicolumn{1}{c}{0.364} &  \multicolumn{1}{c}{0.376} & 	0.356 &	0.36 &	0.38 &	0.37\\
        &\multicolumn{1}{l}{Tulu V2 13B} & \multicolumn{1}{c}{0.376} &  \multicolumn{1}{c}{0.389} & 	0.385 &	0.387 &	0.405 &	0.392\\

        
        &\multicolumn{1}{l}{Orca 2 7B} & \multicolumn{1}{c}{0.44} &  \multicolumn{1}{c}{0.436} & 0.433 &	0.359 &	0.351 &	0.347\\
        &\multicolumn{1}{l}{Orca 2 13B} & \multicolumn{1}{c}{0.445} &  \multicolumn{1}{c}{0.442} & \multicolumn{1}{c}{0.447}	& \multicolumn{1}{c}{0.347} & \multicolumn{1}{c}{0.366} & \multicolumn{1}{c}{0.356}\\
        
        
        &\multicolumn{1}{l}{Stable Beluga 7B} & \multicolumn{1}{c}{0.371} &  \multicolumn{1}{c}{0.395} & 0.407 & 0.377 &	0.398 &	0.396\\
        &\multicolumn{1}{l}{Stable Beluga 13B} & \multicolumn{1}{c}{0.388} &  \multicolumn{1}{c}{	0.394} & 	0.404 &	0.4 &	0.405 &	0.412\\
        
         \hline
         
         \multirow{2}{*}{\textbf{RLHF-tuned}} & \multicolumn{1}{l}{Llama 2 7B Chat} & \multicolumn{1}{c}{0.383} &  \multicolumn{1}{c}{	0.391} & 	0.362&	\textbf{0.333}	& 0.349	& \textbf{0.338}\\
        &\multicolumn{1}{l}{Llama 2 13B Chat} & \multicolumn{1}{c}{0.36} &  \multicolumn{1}{c}{0.393}& 	0.383 &	0.341 &	0.365	& 0.357\\
        \hline
        
         \multirow{4}{*}{\textbf{DPO-tuned}} & \multicolumn{1}{l}{Tulu V2 DPO 7B} & \multicolumn{1}{c}{\textbf{0.325}} &  \multicolumn{1}{c}{\textbf{0.345}} & 0.356 &	0.338 &	0.355 & 0.348\\
        & \multicolumn{1}{l}{Tulu V2 DPO 13B} & \multicolumn{1}{c}{0.359} &  \multicolumn{1}{c}{	\textbf{0.345}} & 	0.385 &	0.37 &	0.383 &	0.368\\
        &\multicolumn{1}{l}{Zephyr 7B $\alpha$}  & \multicolumn{1}{c}{0.360} &  \multicolumn{1}{c}{	0.351} & 	0.357	& 0.343 &	0.352	& 0.353\\
        &\multicolumn{1}{l}{Zephyr 7B $\beta$} & \multicolumn{1}{c}{0.384} &  \multicolumn{1}{c}{	0.359} & 	0.369	& 0.35 &	0.356 &	0.345\\
\hline
\end{tabular}
}
\caption{\textbf{\copernicus\ Probe Results}: Master table for all comparative analysis including the detection of the nine potential paradoxes that can arise due to the three probes outlined in Section \ref{sec:copernicus-framework}; EGISES-JSD is used for the comparative evaluation (lower is better); Table \ref{tab:paradox table}, a summary of the observed paradoxes, is derived from this table. Evaluation Script: \url{https://github.com/KDM-LAB/iCOPERNICUS-EMNLP24}}
\label{tab:Copernicus evaluation}
\vspace{-3mm}
\end{table*}

\subsubsection{Baseline Summarization Models} \label{sec:baseline models}
To understand whether the probed LLMs are superior to specialized PLMs trained on personalized summarization tasks, we examine the same ten SOTA summarization models as in \citep{egises} for comparative benchmarking. Five of them are specifically trained personalized models that follow the PENS framework \citep{pens}: (i) PENS-NRMS Injection-Type 1 (PENS-NRMS T1), (ii) PENS-NRMS Injection-Type 2 (PENS-NRMS T2), (iii) PENS-NAML T1, (iv) PENS-EBNR T1, and (v) PENS-EBNR T2. These models encode the document article using the Transformer encoder \cite{transformer}, deep-neural-model-based user history encoders \cite{ebnr,naml,nrms}, and a Pointer-generator-network-based \cite{pointer-gen} decoder for generating the personalized summaries. The other five non-personalized models are generic SOTA summarizers -- BRIO \citep{brio}, SimCLS \citep{simcls}, BigBird-Pegasus \citep{bigbird}, ProphetNet \citep{prophetnet}, and T5-base \citep{t5-4}. These models were evaluated by providing documents enriched with headlines (reference summaries), serving as cues \citep{egises}.  Since the baseline models \textit{are incapable of ICPL}, \textit{\textbf{\underline{\copernicus\ tests are inapplicable}}} for them, and EGISES-based evaluation is sufficient. The model descriptions are in Appendix \ref{app:baselinemodeldesc}. 

\subsubsection{Hyperparameter Selection}
We conduct temperature ablation within the interval $[0.5, 0.75]$ to balance accuracy and diversity for \underline{personalized} summarization and \textbf{observe almost similar results} to the selected $0.6$.\footnote{See Appendix \ref{app:inference setup} for inference environment (LLM settings and compute resources).} Better ICPL performance might be seen for specific models under more comprehensive ablation, but finding an optimal configuration that generalizes for all LLMs is hard. Nevertheless, the current evaluation is a \textit{useful indication of \textbf{potential paradoxes} and the \textbf{possibility of misguided model selection} if one relies solely on an EGISES-based leaderboard} (see section \ref{sec: EGISES misleading} for empirical results).\footnote{In a way, \copernicus\ tests show the need for detailed hyperparameter optimization before model selection.} 

\section{Observations and Insights}
\subsection{Effect of Examples (User-Summaries)}
In the probe 1, we find that \textbf{10 model variants (out of 17) exhibit an increase in ICPL} (i.e., EGISES-JSD scores) by an average of 2.6\%$\uparrow$ for 2-shot prompts (w/o reading history (hist.)) w.r.t zero-shot (see PX-1 col. of Table \ref{tab:paradox table} for the performing models (denoted as: \textcolor{teal}{\xmark})). However, seven models degrade with (plain) 2-shot (w/o history) prompts (average drop of 1.7\%$\downarrow$), leading to the first of the five observed \textit{\textbf{paradoxes of less is more}} as outlined in section \ref{sec:copernicus-framework} (see Table \ref{tab:paradox table} for result summary). 

\paragraph{(PX-1) Implicit is more than explicit:} We believe that these seven models learn more from the latent concept association in the (temporal) reader's history at a broader level than the specific concepts within the example summaries (i.e., explicit reader-profile as "\textit{writing-history}"), the replacement of which makes them deviate from their earlier ICPL performance. \textit{However, all these models show significant ICPL boost w.r.t their respective base models for the 2-shot w/o hist. case} (denoted by \textbf{\textcolor{teal}{\dag}}). A real example of PX-1 can be seen in Figure \ref{fig:Example PX-1}. 

\begin{table}[t]\normalsize
\centering
\scalebox{0.66}
{
    \begin{tabular}{llcccccc}
    \hline
    & \multicolumn{1}{l}{ \textbf{Model Variants}} & \multicolumn{1}{c}{\textbf{PX-1}} & \multicolumn{1}{c}{\textbf{PX-2}} & \multicolumn{1}{c}{\textbf{PX-3}} & \multicolumn{1}{c}{\textbf{PX-4}} & \multicolumn{1}{c}{\textbf{PX-5}}\\ \hline
            \multirow{2}{*}{\textbf{Base-}} & \multicolumn{1}{l}{Llama 2 7B} & \multicolumn{1}{c}{\textcolor{teal}{\textbf{ {\xmark}}}} & \multicolumn{1}{c}{\textcolor{teal}{\textbf{ {\xmark}}}} &  \multicolumn{1}{c}{\textcolor{red}{\textbf{ {\cmark}}}} &	\textcolor{red}{\textbf{ {\cmark}}} &	\textcolor{red}{\textbf{ {\cmark}}}\\ 
            
            \multirow{2}{*}{\textbf{Model}} & \multicolumn{1}{l}{Llama 2 13B} & \multicolumn{1}{c}{\textcolor{teal}{\textbf{ {\xmark}}}} & \multicolumn{1}{c}{\textcolor{teal}{\textbf{ {\xmark}}}} &  \multicolumn{1}{c}{\textcolor{red}{\textbf{ {\cmark}}}} &	\textcolor{red}{\textbf{ {\cmark}}} &	\textcolor{red}{\textbf{ {\cmark}}}\\ 
            & \multicolumn{1}{l}{Mistral 7B v0.1} & \multicolumn{1}{c}{\textcolor{teal}{\textbf{ {\xmark}}}} & \multicolumn{1}{c}{\textcolor{teal}{\textbf{ {\xmark}}}} &  \multicolumn{1}{c}{\textcolor{red}{\textbf{ {\cmark}}}} & 	\textcolor{red}{\textbf{ {\cmark}}} & \textcolor{red}{\textbf{{\cmark}}}\\
            \hline
            \multirow{7}{*}{\textbf{Instruct-}} & \multicolumn{1}{l}{Mistral 7B Instr. v0.1} & \multicolumn{1}{c}{\textcolor{teal}{\textbf{ {\xmark}}}} & \multicolumn{1}{c}{\textcolor{teal}{\textbf{ {\xmark}}}} &  \multicolumn{1}{c}{\textcolor{teal}{\textbf{ {\xmark}}}} &	\textcolor{red}{\textbf{ {\cmark}}} &	\textcolor{red}{\textbf{ {\cmark}}}\\ 
            & \multicolumn{1}{l}{Mistral 7B Instr. v0.2} & \multicolumn{1}{c}{\textcolor{teal}{\textbf{ {\xmark}}}} & \multicolumn{1}{c}{\textcolor{teal}{\textbf{ {\xmark}}}} &  \multicolumn{1}{c}{\textcolor{teal}{\textbf{ {\xmark}}}} &	{\textcolor{teal}{\textbf{ {\xmark}}}} &	\textcolor{red}{\textbf{ {\cmark}}}\\ 
    
            
            & \multicolumn{1}{l}{Tulu V2 7B} & \multicolumn{1}{c}{\textbf{\textcolor{red}{\cmark}\textcolor{teal}{\dag}}} & \multicolumn{1}{c}{\textcolor{teal}{\textbf{ {\xmark}}}} &  \multicolumn{1}{c}{\textcolor{teal}{\textbf{ {\xmark}}}} &	\textcolor{red}{\textbf{ {\cmark}}} &	\textcolor{red}{\textbf{\cmark}}\\ 
            & \multicolumn{1}{l}{Tulu V2 13B} & \multicolumn{1}{c}{\textbf{\textcolor{red}{\cmark}\textcolor{teal}{\dag}}} & \multicolumn{1}{c}{\textcolor{red}{\textbf{ {\cmark}}}} &  \multicolumn{1}{c}{\textcolor{red}{\textbf{ {\cmark}}}} &	\textcolor{red}{\textbf{ {\cmark}}} &	\textcolor{red}{\textbf{ {\cmark}}}\\ 
    
            
            \multirow{1}{*}{\textbf{tuned}} & \multicolumn{1}{l}{Orca 2 7B} & \multicolumn{1}{c}{\textcolor{teal}{\textbf{ {\xmark}}}} & \multicolumn{1}{c}{\textcolor{teal}{\textbf{ {\xmark}}}} &  \multicolumn{1}{c}{\textcolor{teal}{\textbf{ {\xmark}}}} &	{\textcolor{teal}{\textbf{ {\xmark}}}} &	{\textcolor{teal}{\textbf{ {\xmark}}}}\\ 
            & \multicolumn{1}{l}{Orca 2 13B} & \multicolumn{1}{c}{\textcolor{teal}{\textbf{ {\xmark}}}} & \multicolumn{1}{c}{\textcolor{red}{\textbf{ {\cmark}}}} &  \multicolumn{1}{c}{\textcolor{teal}{\textbf{ {\xmark}}}} &	{\textcolor{teal}{\textbf{ {\xmark}}}} &	\textcolor{red}{\textbf{ {\cmark}}}\\ 
            
            & \multicolumn{1}{l}{Stable Beluga 7B} & \multicolumn{1}{c}{\textbf{\textcolor{red}{\cmark}\textcolor{teal}{\dag}}} & \multicolumn{1}{c}{\textcolor{red}{\textbf{ {\cmark}}}} &  \multicolumn{1}{c}{\textcolor{red}{\textbf{ {\cmark}}}} &	\textcolor{red}{\textbf{ {\cmark}}} &	\textcolor{red}{\textbf{ {\cmark}}}\\ 
            & \multicolumn{1}{l}{Stable Beluga 13B} & \multicolumn{1}{c}{\textbf{\textcolor{red}{\cmark}\textcolor{teal}{\dag}}} & \multicolumn{1}{c}{\textcolor{red}{\textbf{ {\cmark}}}} &  \multicolumn{1}{c}{\textcolor{red}{\textbf{ {\cmark}}}} &	\textcolor{red}{\textbf{ {\cmark}}} &	\textcolor{red}{\textbf{ {\cmark}}}\\ 
            
             \hline
             
            \multirow{1}{*}{\textbf{RLHF-}}& \multicolumn{1}{l}{Llama 2 7B Chat} & \multicolumn{1}{c}{\textbf{\textcolor{red}{\cmark}\textcolor{teal}{\dag}}} & \multicolumn{1}{c}{\textcolor{teal}{\textbf{ {\xmark}}}} &  \multicolumn{1}{c}{\textcolor{teal}{\textbf{ {\xmark}}}} &	\textcolor{teal}{\textbf{ {\xmark}}} &	\textcolor{red}{\textbf{ {\cmark}}}\\ 
            \multirow{1}{*}{\textbf{tuned}}& \multicolumn{1}{l}{Llama 2 13B Chat} & \multicolumn{1}{c}{\textbf{\textcolor{red}{\cmark}\textcolor{teal}{\dag}}} & \multicolumn{1}{c}{\textcolor{red}{\textbf{ {\cmark}}}} &  \multicolumn{1}{c}{\textcolor{teal}{\textbf{ {\xmark}}}} &	\textcolor{teal}{\textbf{ {\xmark}}} &	\textcolor{red}{\textbf{ {\cmark}}}\\ 
            \hline

            \multirow{3}{*}{\textbf{DPO-}}& \multicolumn{1}{l}{Tulu V2 DPO 7B} & \multicolumn{1}{c}{\textbf{\textcolor{red}{\cmark}\textcolor{teal}{\dag}}} & \multicolumn{1}{c}{\textcolor{red}{\textbf{ {\cmark}}}} &  \multicolumn{1}{c}{\textcolor{red}{\textbf{ {\cmark}}}} &	\textcolor{red}{\textbf{ {\cmark}}} &	\textcolor{red}{\textbf{ {\cmark}}}\\ 
            & \multicolumn{1}{l}{Tulu V2 DPO 13B} & \multicolumn{1}{c}{\textcolor{teal}{\textbf{ {\xmark}}}} & \multicolumn{1}{c}{\textcolor{red}{\textbf{ {\cmark}}}} &  \multicolumn{1}{c}{\textcolor{red}{\textbf{ {\cmark}}}} &	\textcolor{red}{\textbf{ {\cmark}}} &	\textcolor{teal}{\textbf{ {\xmark}}}\\ 
    
            
            \multirow{1}{*}{\textbf{tuned}}& \multirow{1}{*}{Zephyr 7B $\alpha$} & \multicolumn{1}{c}{\textcolor{teal}{\textbf{ {\xmark}}}} & \multicolumn{1}{c}{\textcolor{teal}{\textbf{ {\xmark}}}} &  \multicolumn{1}{c}{\textcolor{teal}{\textbf{ {\xmark}}}} &	\textcolor{red}{\textbf{ {\cmark}}} &	\textcolor{red}{\textbf{ {\cmark}}}\\ 
            & \multirow{1}{*}{Zephyr 7B $\beta$} & \multicolumn{1}{c}{\textcolor{teal}{\textbf{ {\xmark}}}} & \multicolumn{1}{c}{\textcolor{teal}{\textbf{ {\xmark}}}} &  \multicolumn{1}{c}{\textcolor{teal}{\textbf{ {\xmark}}}} &	\textcolor{teal}{\textbf{ {\xmark}}} &	\textcolor{teal}{\textbf{ {\xmark}}}\\ 
            
    \hline
    \end{tabular}
}
\caption{\textbf{Paradox (PX) of less is more (\textcolor{red}{\cmark}: PX exists):} PX-1/2: 2-shot w/o \& w/ hist.; PX-3/4/5: C-0-shot/C-2-shot w/o \& w/ hist.; \textbf{\textcolor{teal}{\dag}} denotes improvement over base models; for examples see Figures \ref{fig:Example PX-1}-\ref{fig:Ex_PX5} in Appendix.}
\label{tab:paradox table}
\vspace{-5mm}
\end{table}


\begin{table*}[th]
\centering
\scalebox{0.7}{
\begin{tabular}{lc}
\hline
\textbf{Baseline Models} & \textbf{EGISES-JSD} \\
\hline
\textbf{BigBird-Pegasus} & \textbf{0.429}\\
\textbf{SimCLS} & 0.557  \\
\textbf{BRIO} & 0.661 \\ \hdashline
\textbf{PENS-NAML T1} & \underline{0.899} \\
\textbf{PENS-NRMS T1} & \underline{0.916}  \\
\hline
\end{tabular}
\begin{tabular}{lcc}
\hline
\textbf{LLM Models} & \textbf{EGISES-JSD} & \textbf{Paradoxes (PX) Observed} \\
\hline
\textbf{Tulu V2 DPO 7B} (0-shot / C-0-shot / 2-shot) & {\textbf{0.325}} / 0.338 / 0.345 & PX-1/2/3/4/5\\
\textbf{Llama 2 7B Chat} (C-0-shot / C-2-shot / 2-shot) & 0.333 / 0.338 / 0.345 & PX-1/5 \\  
\textbf{Llama 2 13B Chat} (C-0-shot) & 0.341 & PX-1/2/5\\
\textbf{Tulu V2 DPO 13B} (2-shot) & 0.345 & PX-2/3/4\\
\textbf{Orca 2 13B} (C-0-shot) & 0.347 & PX-2/5 \\
\hline
\end{tabular}
}
\caption{\textbf{EGISES Leaderboard Misleads}: Top-5 LLMs as per EGISES-JSD, apparently beating top-5 baselines (\underline{top-2 personalized} and top-3 non-personalized w/ summary cue), do not pass critical \copernicus\ tests exhibiting several paradoxes.}
\label{tab:ICPL apparently exists}
\vspace{-5mm}
\end{table*}

\subsection{Effect of User's Reading History}
As a part of the second probe w.r.t \copernicus, we observe that \textbf{10 model variants exhibit an increase in ICPL} (avg. boost: 2.5\%$\uparrow$) for 2-shot w/ hist. w.r.t zero-shot. It is also observed that the set of LLMs that show ICPL in the previous case does not take advantage of the additional history data, leading to the second of the paradoxes (i.e., PX-2). 
\paragraph{(PX-2) Reading history distracts:} Seven model variants show worse ICPL with the additional history data (avg. drop: 2\%$\downarrow$; see Table \ref{tab:paradox table}). We believe that these models tend to learn more from the format of the prompt and latent-concept association but at a rather broader thematic level and get "\textit{distracted}" \cite{llm-distraction} by the concept distribution within histories. A real example of PX-2 can be seen in Figure \ref{fig:Ex_PX2}.      

\subsection{Effect of Contrastive Prompts}
The third probe tests if contrastive user profiles (C-0-shot, C-2-shot w/o hist., C-2-shot w/ hist.) induce better ICPL in the models. This probe is central for any model to pass the \copernicus\ test. 

\textbf{Case of C-0-shot}: We observe that \textbf{9 model variants seem to harness the additional contrastive information about the readers' reading history} in comparison with the 0-shot case (avg. boost: 3.8\%$\uparrow$, see PX-3 col. of Table \ref{tab:paradox table} for performing models), which is better than the overall boost observed with (plain) 2-shot (w/ and w/o history). This indicates that these models might actually be utilizing the contrastive information in the reading histories rather than the examples without the contrast. However, this ICPL behavior is not observed in the remaining eight model variants, leading to a special case of the \textbf{paradox of contrastive user profiles} - PX-3. We will discuss this subsequently.

\textbf{Case of C-2-shot w/o history}: We observe that only \textbf{six model variants pass this probe test and seem to harness the contrastive examples \underline{quite notably}} ($p$-value $<0.01$) when compared to the case of 2-shot w/o history (avg. boost: 4.1\%$\uparrow$) (see PX-4 col. in Table \ref{tab:paradox table} for these 6 models). In fact, Llama 2 13B Chat, one of the 6 variants, degrades in the cases of (plain) 2-shot (w \& w/o hist.). Again, most variants (11) are non-compliant, leading to the other special case -- PX-4.

\textbf{Case of C-2-shot w/ history}: We find that \textbf{\underline{only} three model variants are compliant with this probe and seem to \underline{marginally} (i.e., not notable) utilize the additional contrastive history} along with the contrastive examples when compared to the case of C-0-shot (avg. boost: 0.6\%$\uparrow$; Orca 2 7B has max. boost of 1.2\%$\uparrow$), while 14 model variants clearly seem to get distracted with an average drop of 1.6\%$\downarrow$, leading to the paradox -- PX-5.

\paragraph{(PX-3/4/5) Contrast can be confusing:} Surprisingly, additional contrastive reader profile information doesn't enhance ICPL in many model variants. Eight out of 17 models show distraction and a 1.6\% average drop when contrast is injected into the history component, struggling to map intra-concept associations within the history (\textbf{PX-3}). A real example of PX-3 can be seen in Figure \ref{fig:Ex_PX3}. Moreover, 11 models exhibit a 3.6\% average drop when provided with contrastive examples in a 2-shot setup without history (\textbf{PX-4}). This decline may be due to overfitting the format of the 2-shot setup, causing distraction. A real example of PX-4 can be seen in Figure \ref{fig:Ex_PX4}. Additionally, 14 model variants fail to effectively utilize the richest prompt, C-2-shot with history, with seven unable to harness any form of contrastive profile information (\textbf{PX-3/4/5}). Three variants use contrastive history in C-0-shot but not contrastive examples in C-2-shot w/o hist., thus learning from broader historical topics rather than label instances (\textbf{PX-4/5}). Surprisingly, four variants can utilize contrastive histories in C-0-shot and contrastive examples in C-2-shot w/o hist., but not both, potentially due to spurious (and cross) linking between history and example concepts. A real example of PX-5 can be seen in Figure \ref{fig:Ex_PX5}.

\subsection{Effect of Article Length}
In probe 1, we substitute longer articles with shorter ones, keeping the prompt length the same (see Table \ref{tab:ContextLengthTable}). Arguably, personalized summarization of shorter articles should be an easier task. However, several models exhibit the PX-1 and PX-1 (contrast) paradoxes (i.e., adding examples led to poor ICPL) under probe-1. Hence, we \textit{\textbf{cannot generalize} that the length of the articles has influence}. Also, several models pass tests involving longer articles or history sequences, thereby reinforcing this finding.

\subsection{EGISES Leaderboard Misleads}\label{sec: EGISES misleading}
We observe from Table \ref{tab:paradox table} that the top-5 best-performing LLMs in terms of EGISES-JSD as per Table \ref{tab:ICPL apparently exists}\footnote{All the models are \textbf{\underline{\textit{seemingly}} better} than the strongest baseline model (BigBird-Pegasus), with 57\%$\uparrow$ for Tulu V2 DPO 7B when compared to PENS-NAML-T1.} - i.e., Tulu v2 DPO 7B, Llama 2 7B Chat, Llama 2 13 B Chat, Tulu v2 DPO 7B, and Orca 2 13 B show several paradoxes with Tulu v2 DPO 7B being the worst. In fact, the best performing models (Orca 2 7B and Zephyr 7B $\beta$), passing all the \copernicus\ tests (see Table \ref{tab:paradox table}) do not rank high in Table \ref{tab:ICPL apparently exists}. This empirically establishes that \textit{EGISES as a standalone measure is \textbf{inadequate for ICPL evaluation} in LLMs}. 

\section{Validation of Paradoxes}\label{sec: Paradox Validation}
The paradoxes raise a serious question: \textit{do these models exhibit true ICPL?} (including the top-5 LLMs in Table \ref{tab:ICPL apparently exists}) Hence, we need to first confirm that the paradoxes exist, which is done via adversarial probes. We select three model variants for three scenarios: (i) \textbf{worst case}: Tulu V2 DPO 7B showing all the paradoxes, (ii) \textbf{average case}: Llama 2 13B Chat showing non-contrastive (PX-1/2) and contrastive (PX-5) paradoxes, and (iii) \textbf{best case}: Mistral 7B Instruct v0.2 having PX-5 only.  

\begin{table}[t]
    \centering
    \scalebox{0.8}{
    \begin{tabular}{lccc}
        \toprule
        \hline
        \textbf{Models}&\textbf{PX-3}&\textbf{PX-4} &\textbf{PX-5}\\
        \midrule
        \hline
        Llama 2 13B Chat& 1.97\% \textcolor{red}{$\uparrow$}& 0.28\% \textcolor{red}{$\uparrow$}& 1.54\% \textcolor{red}{$\uparrow$}\\
        
        Mistral 7B Inst. v0.2& 1.96\% \textcolor{red}{$\uparrow$}& 0.46\% \textcolor{red}{$\uparrow$}& 1.78\% \textcolor{red}{$\uparrow$}\\
        
        Tulu V2 DPO 7B& 1.94\% \textcolor{red}{$\uparrow$}& 0.56\% \textcolor{red}{$\uparrow$}& 2.2\% \textcolor{red}{$\uparrow$}\\
        \hline
    \end{tabular}
    }
    \caption{\textbf{Adversarial Validation:} PX-3/4/5 exists.\textbf{\textcolor{red}{\%$\uparrow$}} value indicates how much the paradox worsens.}
    \label{tab:Adversarial validation}
    \vspace{-5mm}
\end{table}

\textbf{Adversarial Probe-based Validation:} In the adversarial probe setup, one of the user profiles (i.e., ground-truth history or examples) in the contrastive prompts (C-0-shot/C-2-shot w/o hist./C-2-shot w/ hist.) was replaced with a random history/examples. Since it is randomly sampled, the noise would be completely irrelevant to the document article to be summarized. Upon injection of such a noise, a model's personalization performance, and thereby the exhibited paradoxes, should degrade further. However, \textit{if a model shows a better or similar EGISES score, then it would mean that the contrastive tests of iCOPERNICUS are no better than tests based on random choices}. In other words, \copernicus\ probes may not necessarily provide conclusive insights. However, we observe in Table \ref{tab:Adversarial validation} that the paradoxes do worsen (PX-3 avg. spike: 1.9\%\textcolor{red}{$\uparrow$}; PX-4 avg. spike: 0.43\%\textcolor{red}{$\uparrow$}; PX-5 avg. spike: 1.8\%\textcolor{red}{$\uparrow$}). \textbf{This validates the robustness of the \copernicus\ tests}. Details in Appendix \ref{app:adversarial test results}. 

\textbf{Human-Judgment Validation:} We further validated PX-5 (the most serious paradox) using survey-based unbiased judgments (i.e., similarity ratings (1 (low) - 6 (very high)) on summary-pairs, both reference and model-generated from 339 respondents.\footnote{Grad student volunteers in 20-30 age group from Computer Sc., Maths, and Humanities; $\sim$70\% male, $\sim$30\% female. For details of survey methodology see Appendix \ref{app:HJ Validation}.} The core objective is to validate the extent to which human evaluators would \textit{agree with the design principles of EGISES-JSD at a cognitive level} and thereby agree with the \copernicus\ test results that are based on EGISES-JSD. We, therefore, model a "\textit{human version}" of EGISES-JSD (i.e., EGISES-HJ) and used the survey responses to estimate EGISES-HJ). We find that PX-5 persists except for Mistral 7B Instruct v0.2 (best case; ICPL boost: 1.42\%\textcolor{teal}{$\uparrow$}; Table \ref{tab:HJ Validation}). We also find that PX-1 (contrastive) persists in the other models.\footnote{Statistical Significance: $p$-value $< 0.01$.}

\section{Related Work}
\textbf{ICL for summarization:}
ICL capabilities in LLMs w.r.t summarization were first observed within reinforcement learning frameworks, utilizing human-feedback-trained reward models to predict human ratings for specific summary actions \citep{openai,reward-model-1}. Since then, LLMs have shown unprecedented ICL-based summarization performance \citep{ICL-summarization-news,ICL-summarization-meeting,ICL-summarization-dialogue}. This opens the possibility of ICPL in these LLMs. At the same time, it also underscores the necessity for robust and dependable methods of evaluating the degree of ICPL within such models. Benchmarking LLMs for summarization was done for accuracy, fluency, and consistency \citep{LLM_Sum_Bench}, but not ICPL. 

\textbf{Personalization evaluation:} 
Personalization evaluation is studied in recommendation systems \citep{recsys-eval-survey}, with metrics based on the Jaccard Index, MAE/RMSE/Hit-Ratio \citep{recsyseval-1}, and nDCG (normalized Discounted Cumulative Gain) \citep{personalizationmeasure2}. However, these metrics are not useful for summarization. The only personalization evaluation metric for summarizers is EGISES\ \citep{egises}. However, as established, relying solely on EGISES to evaluate ICPL can be misleading. 

\section{Conclusion}
We propose \copernicus, a novel evaluation framework for analyzing the capability of true In-Context Personalization Learning (ICPL) in LLMs for the personalized summarization task. The central goal is to detect whether models can pass all the probes (or exhibit the "\textit{paradox of less is more}"). We showed that relying solely on EGISES-scores can be misleading (as in top-4 LLMs beating baselines). Only 2 out of the 17 SOTA LLM variants probed passed the \copernicus\ test, hence the need for further research on ICPL-driven LLMs.   


\section*{Limitations}
In this work, we restrict probing of In-Context Personalization Learning (ICPL) w.r.t summarization to  7B and 13B model variants. It would be interesting to observe how smaller SLMs ($<$ 7B), such as the Phi model suite \citep{phi}, fare the \copernicus\ test. Also, studies around the effect of more recent fine-tuning techniques such as PEFT (LoRA and QLoRA) need to be analyzed. On similar lines, it would be interesting to observe whether ICPL, which otherwise is not emerging by doubling the models' size, finally starts emerging at an even larger scale (which often is the case for several emerging properties). At the same time, robust and reliable ICPL measures should be designed for aggregated leaderboard generation of models w.r.t ICPL within the \copernicus\ framework. Finally, more robust and systematic adversarial probing of ICPL is required to analyze true ICPL in models.

\section*{Ethics Statement}
We would like to declare that we used the PENS dataset prepared and released by Microsoft Research. Our human-judgment survey was conducted according to the norms set by the Institutional Review Board (IRB) and respects participant anonymity as per guidelines.

\bibliography{ICPL}
\bibliographystyle{acl_natbib}

\appendix
\section{Preliminaries}
\subsection{Degree-of-personalization}\label{app:insensitivity}
In this section, we recall the notion of \textit{insensitivity-to-subjectivity} of a personalized summarization model ($M_{\boldsymbol{\theta},u}: (d,u) \mapsto s_u; \text{where }s_{u}$ is the personalized summary for reader $u$ on document $d$) as defined in \citet{egises}.

\begin{definition} \label{def:weak insensitivity}
    \textbf{Weak Insensitivity-to-Subjectivity.} A summarization model $M_{\boldsymbol{\theta},u}$ is (weakly) Insensitive-to-Subjectivity w.r.t reader $u$, if $\forall(u_{i}, u_{j})$, $(\sigma(u_{i}, u_{j}) \leq \tau_{max}^{U}) \iff
(\sigma(s_{u_i}, s_{u_j}) > \tau_{max}^{S})$, 
where $\sigma$\footnote{$\sigma( u_{i},  u_{i}) = 0$ ; $\sigma(u_{i},  u_{j}) \in [0, 1]$; $\sigma$ satisfies positivity, reflexive, maximality, symmetry, and the triangle inequality.} is an arbitrary distance metric defined on the metric space $\mathcal{M}$ where $d, u, s$ are defined, $\tau_{max}^{U}$ is the maximum limit for $u_{i},  u_{j}$ to be mutually indistinguishable, and $\tau_{max}^{S}$ is the maximum limit for $s_{u_i},  s_{u_j}$ to be mutually indistinguishable.
\end{definition}

\begin{definition} \label{def:strong insensitity}
    \textbf{Strong Insensitivity-to-Subjectivity.} A summarization model $M_{\boldsymbol{\theta},u}$ is (strongly) Insensitive-to-Subjectivity w.r.t reader $u$
    if $\forall(u_{i}, u_{j})$, $M_{\boldsymbol{\theta},u}$ satisfies: (i) the condition of weak insensitivity, and (ii) $(\sigma(u_{i}, u_{j}) > \tau_{max}^{U}) \iff (\sigma(s_{u_i}, s_{u_j}) \leq \tau_{max}^{S})$.
\end{definition}

\subsection{EGISES and Degree-of-Personalization} \label{app: egises}
We generalize the definition of summary-level deviation (or, \textit{\textbf{Deg}ree-of-\textbf{R}esponsiven\textbf{ess}} (\degress))\footnote{This is based on the notion of weak and strong \textit{insensitivity-to-subjectivity}, as defined by \cite{egises} (see Appendix \ref{app:insensitivity}).} proposed by \citet{egises} as follows: 

\begin{definition} \label{def:responsiveness}
    \textbf{Summary-level \degress.} Given a document $d_i$ and a user-profile $u_{ij}$ (user $j$'s expected summary), the summary-level responsiveness of a personalized model $M_{\boldsymbol{\theta},u}$, (i.e., $\degress(s_{u_{ij}}|(d_{i}, u_{ij}))$), is defined as the \textit{proportional} divergence between model-generated summary $s_{u_{ij}}$ of $d_i$ for $j$-th user from other user-specific summary versions w.r.t a corresponding divergence of $u_{ij}$ from the other user-profiles. 
\end{definition}

$\degress(s_{u_{ij}}|(d_{i}, u_{ij}))$ is formulated as: 
\begin{equation} \label{eq:degress}\small
\begin{split}
        &\degress(s_{u_{ij}}|(d_{i}, u_{ij})) = \frac{1}{|\textbf{U}_{d_i}|} \sum\limits_{k=1}^{|\textbf{U}_{d_i}|} \frac{min(X_{ijk}, Y_{ijk})+\epsilon}{max(X_{ijk},Y_{ijk})+\epsilon}\\
        &X_{ijk} = \frac{\exp(w(u_{ij} | u_{ik}))}{\sum\limits_{l=1}^{|\textbf{U}_{d_{i}}|}{\exp(w(u_{ij} | u_{il}))}} \cdot \sigma(u_{ij} , u_{ik})\\
        &Y_{ijk} = \frac{\exp(w(s_{u_{ij}} | s_{u_{ik}}))}{\sum\limits_{l=1}^{|\textbf{U}_{d_{i}}|}{\exp(w(s_{u_{ij}} | s_{u_{il}}))}} \cdot \sigma(s_{u_{ij}} , s_{u_{ik}})\\
        &{w(u_{ij} | u_{ik})  = \frac{\sigma(u_{ij} , u_{ik})}{\sigma(u_{ij} , d_{i})}}; \ \ {w({s_{u_{ij}}} | s_{u_{ik}})  = \frac{\sigma(s_{u_{ij}} , s_{u_{ik}})}{\sigma(s_{u_{ij}} , d_{i})}}
\end{split}
\end{equation}  
Here, $|\textbf{D}|$ is the total number of documents in the evaluation dataset, $|\textbf{U}|$ is the total number of users who created gold-reference summaries that reflect their expected summaries (and thereby their subjective preferences or profiles), and $|\textbf{U}_{d_i}|$ ($=|\textbf{S}_{d_i}|$) is the number of users who created gold-references for document $d_i$. A lower value of $\degress(s_{u_{ij}}|(d_{i}, u_{ij}))$ indicates that while reader-profiles are different, the generated summary $s_{u_{ij}}$ is very similar to other reader-specific summaries (or vice versa), and hence, is not responsive at the summary-level. The system-level \degress \space and EGISES have been formulated as follows:
\begin{equation}\label{eq:Degree-of-Responsiveness}\small
\degress(M_{\boldsymbol{\theta},u}) = \frac{\sum\limits_{i=1}^{|\textbf{D}|}\frac{\sum\limits_{j=1}^{|\textbf{U}_{d_i}|} \degress(s_{u_{ij}}|(d_{i}, u_{ij}))}{|\textbf{U}_{d_i}|}}{|\textbf{D}|}
\end{equation}

\begin{equation} \label{eq:system-level-egises}\small
    \egises(M_{\boldsymbol{\theta},u}) = 1 - \degress(M_{\boldsymbol{\theta},u})
\end{equation}

EGISES measures the degree of insensitivity-to-subjectivity for relative benchmarking of how much models lack personalization (i.e., a lower score is better within the range: $[0,1]$) instead of assigning an absolute goodness score. In this paper, we choose Jensen-Shannon Divergence (JSD) \cite{jsd}, where $d$, $u$, and $s_{u}$ are defined as word distributions on a probability space. JSD has a strong human-judgment correlation and has been used in evaluating the ten specialized baseline models \citep{egises}.

\subsection{In-Context Learning}\label{app: ICL description}
ICL is a method employed by LLMs, notably emphasized in \citet{ICL-few-shot-GPT3} (GPT-3's ICL behavior was first highlighted), where models acquire proficiency in apparently unknown tasks (i.e., tasks on which the models are not pre-trained) from limited examples, called {\em prompts}, with no update in their parameters (i.e., the models are frozen). Formally, it is defined below.

\begin{definition} \label{def:prompt}
    \textbf{Prompt}: A prompt $\mathcal{P}$ given to a language model $M$ is a sequence of $n$ concatenated ($\oplus$) demonstration examples (i.e., input-label pairs: 
    $\left(x_{i}^{\prime}\oplus y_{x_i}^{\prime}\right)$) as $\langle \left(x_{1}^{\prime}\oplus y_{x_1}^{\prime}\right) \oplus\left(x_{2}^{\prime}\oplus y_{x_2}^{\prime}\right) \oplus\ldots\oplus\left(x_{n}^{\prime}\oplus y_{x_n}^{\prime}\right) \rangle$ and an input query $x_q$ appended, such that $x_i \neq x_q$. 
\end{definition}

The input often includes a description of the task or a system command $\mathcal{T}$ before the demonstration sequence $\mathcal{D}_{\mathcal{P}}$. 
We now provide a formal definition of ICL as follows:

\begin{definition}\label{def: ICL}
    \textbf{In-Context Learning (ICL):} A model $M$ is said to exhibit ICL if given a prompt $\mathcal{P}$ $\sim \mathbb{D}$ (where $\mathbb{D}$ is an unseen demonstration dataset) and an unseen task $\mathcal{T}$, $M:(\mathcal{T} \oplus \mathcal{D}_{\mathcal{P}}\oplus x_q) \mapsto  y^*_{x_q};\ y^*_{x_q} \in Y^*_{x_q};$ where $Y^*_{x_q}$ is the expected set of output labels for the given query $x_q$.
\end{definition}

$M$ predicts (i.e., maps) using the prompt's conditioning only, requiring it to discern essential aspects such as input-label mapping, input text distribution, label space, and formatting (lexico-syntactic structural relationship between the prompt components).

\subsection{ICL is latent concept mapping}\label{app: latent-concept-learning}
\citet{ICL-latent-concept-learning-1} suggested that LLMs acquire latent document-level concepts during pretraining to generate coherent subsequent tokens. ICL occurs when LLMs identify shared latent concepts among prompt examples. \citet{ICL-latent-concept-learning-2} revealed that input-label mapping, input-text distribution, label space, and format in prompts matter more. \citet{ICL-latent-concept-learning-with_size-4} provided supportive evidence that label association learning becomes more pronounced with larger model sizes. Although further analysis needs to be done, we consider the latent-concept association hypothesis the most compelling explanation of ICL so far and use this as a primary tool for understanding our own findings on ICPL in this paper. In the following section, we first define ICPL (w.r.t personalized summarization) as a special case of ICL and then propose \copernicus\ as a framework for evaluating ICPL.

\section{Model Descriptions}
\subsection{LLM Model Descriptions} \label{app:LLMdesc}
We provide concise descriptions of the LLMs analyzed in this study for understanding ICPL.

\begin{enumerate}
    \item \textbf{Llama 2} - Llama 2 \cite{llama-2-base-underfitting} is a family of transformer-based autoregressive causal language models, ranging in scale from 7 billion to 70 billion parameters. Llama 2 models are trained on 2 trillion tokens and have double the context length of Llama 1.
    
    \item \textbf{Llama 2 Chat} - Llama 2 Chat \cite{llama-2-base-underfitting} is a fine-tuned version of Llama 2, optimized for dialogue applications using reinforcement learning from human feedback (RLHF). Llama 2 Chat models demonstrate improved helpfulness and safety compared to other open models and achieve comparable performance to ChatGPT according to human evaluations.
    
    \item \textbf{Mistral 7B} - Mistral 7B \cite{mistral-(GQA-SWA)} is a language model that outperforms Llama 2 13B and Llama 1 34B in various tasks, such as natural language inference, mathematics, and code generation. It leverages grouped-query attention (GQA), and sliding window attention (SWA). GQA significantly accelerates the inference speed and also reduces the memory requirement during decoding.
    
    \item \textbf{Mistral 7B Instruct} - Mistral 7B Instruct \cite{mistral-(GQA-SWA)} is a fine-tuned version of Mistral 7B that leverages instruction datasets to enhance its generalization and adaptation capabilities. The model has two versions: v0.1 and v0.2. It exhibits superior performance compared to all 7B models on MT-Bench and is comparable to 13B – Chat models. Mistral 7B Instruct v0.2 is an updated version with improvements in instruction following and generalization capabilities.
    
    \item \textbf{Tulu V2 Suite} - The Tulu V2 Suite \cite{Tulu} is a collection of fine-tuned large language models (LLMs) based on Llama 2. The models in this suite are fine-tuned on a mix of publicly available, synthetic, and human datasets. The suite includes Tulu V2 models as well as the DPO fine-tuned Tulu V2 DPO models. 
    
    \item \textbf{Orca 2} - Orca 2 \cite{orca-2} is a suite of models that are fine-tuned on Llama 2 using synthetic dataset. Orca models are designed to enhance the reasoning abilities of smaller language models by imitating the step-by-step reasoning traces of more capable LLMs. Orca 2 models surpass models of similar size and attain performance levels similar to or better than models five times larger on complex reasoning tasks.
    
    \item \textbf{Stable Beluga} - Stable Beluga \cite{stable-beluga-2} is a collection of models that have been fine-tuned on the Llama 2 using an internal Orca style dataset. The primary objective of these models is to generate responses that are not only responsive to user prompts and queries, but also emphasize reasoning and helpfulness.
    
    \item \textbf{Zephyr 7B} - Zephyr 7B \cite{Zephyr} is a series of language models trained to act as helpful assistants developed by the HuggingFace H4 team. This includes Zephyr 7B $\alpha$ and Zephyr 7B $\beta$. It is a fine-tuned on Mistral 7B v0.1 and it was trained on on a mix of publicly available, synthetic datasets using DPO.
    
\end{enumerate}

\subsection{Baseline Model Descriptions} \label{app:baselinemodeldesc}
We briefly introduce the SOTA baseline summarization models that were analyzed to understand their degree-of-personalization below:

\begin{enumerate}
    \item \textbf{PENS-NRMS Injection-Type 1}: The PENS framework \citep{pens} utilizes NRMS (Neural News Recommendation with Multi-Head Self-Attention) \citep{nrms} for personalized summary generation. NRMS employs a news encoder using multi-head self-attention to understand news titles and learn user representations based on browsing history. Additive attention enhances learning by selecting important words and articles. In Injection-Type 1, NRMS user embedding is injected into PENS by initializing the decoder’s hidden state of the headline generator.
    
    \item \textbf{PENS-NRMS Injection-Type 2}: To generate a personalized summary, NRMS user embedding is injected into attention values (Injection-Type 2) of PENS that helps to personalize attentive values of words in the news body.

    \item \textbf{PENS-NAML Injection-Type 1}: NAML (Neural News Recommendation with Attentive Multi-View Learning) \citep{naml} incorporates a news encoder employing a multi-view attention model for comprehensive news representations. The user encoder learns user representations based on interactions with browsed news, allowing the selection of informative news. In Injection-Type 1, this user embedding is injected into the PENS model for personalization.
    
    \item \textbf{PENS-EBNR Injection-Type 1}: EBNR (Embedding-based News Recommendation for Millions of Users) \cite{ebnr} proposes a method for user representations by using an RNN model that takes browsing histories as input sequences. This user embedding is injected using Type 1 into the PENS model for personalization.
    
    \item \textbf{PENS-EBNR Injection-Type 2}: This personalized model injects EBNR user embedding into PENS using type-2. 
    
    \item \textbf{BRIO}: BRIO \cite{brio} assumes a non-deterministic training paradigm that assigns probability mass to different candidate summaries according to their quality, thereby helping it to better distinguish between high-quality and low-quality summaries.
    
    \item \textbf{SimCLS}: SimCLS (A Simple Framework for Contrastive Learning of Abstractive Summarization) \cite{simcls} uses a two-stage training procedure. In the first stage, a Seq2Seq model (BART \citep{bart}) is trained to generate candidate summaries with MLE loss. Then, a RoBERTa-initiated evaluation model is trained to rank these using contrastive learning.
    
    \item \textbf{BigBird-Pegasus}: BigBird \cite{bigbird} is an extension of Transformer based models designed specifically for processing longer sequences. It utilizes sparse, global, and random attention mechanisms to approximate full attention which enables it to handle longer contexts more efficiently. 
    
    \item \textbf{ProphetNet}: ProphetNet \cite{prophetnet} is a seq2seq pre-trained model that employs n-gram prediction using the n-stream self-attention mechanism. It enhances n-step ahead prediction by predicting the next n tokens at once, based on previous tokens, thus avoiding overfitting on local correlations.
    
    \item \textbf{T5}: T5 (Text-To-Text Transfer Transformer) is based on the Transformer Encoder-Decoder architecture that operates on the principle of the unified text-to-text task for any NLP problem, including summarization. See \cite{T5-1,T5-2,T5-3} for recent T5 summarization analyses.
\end{enumerate}

\section{PENS Dataset}\label{app: PENS Dataset details}
\begin{figure}[t]
    \centering
    \includegraphics[width=0.5\textwidth]{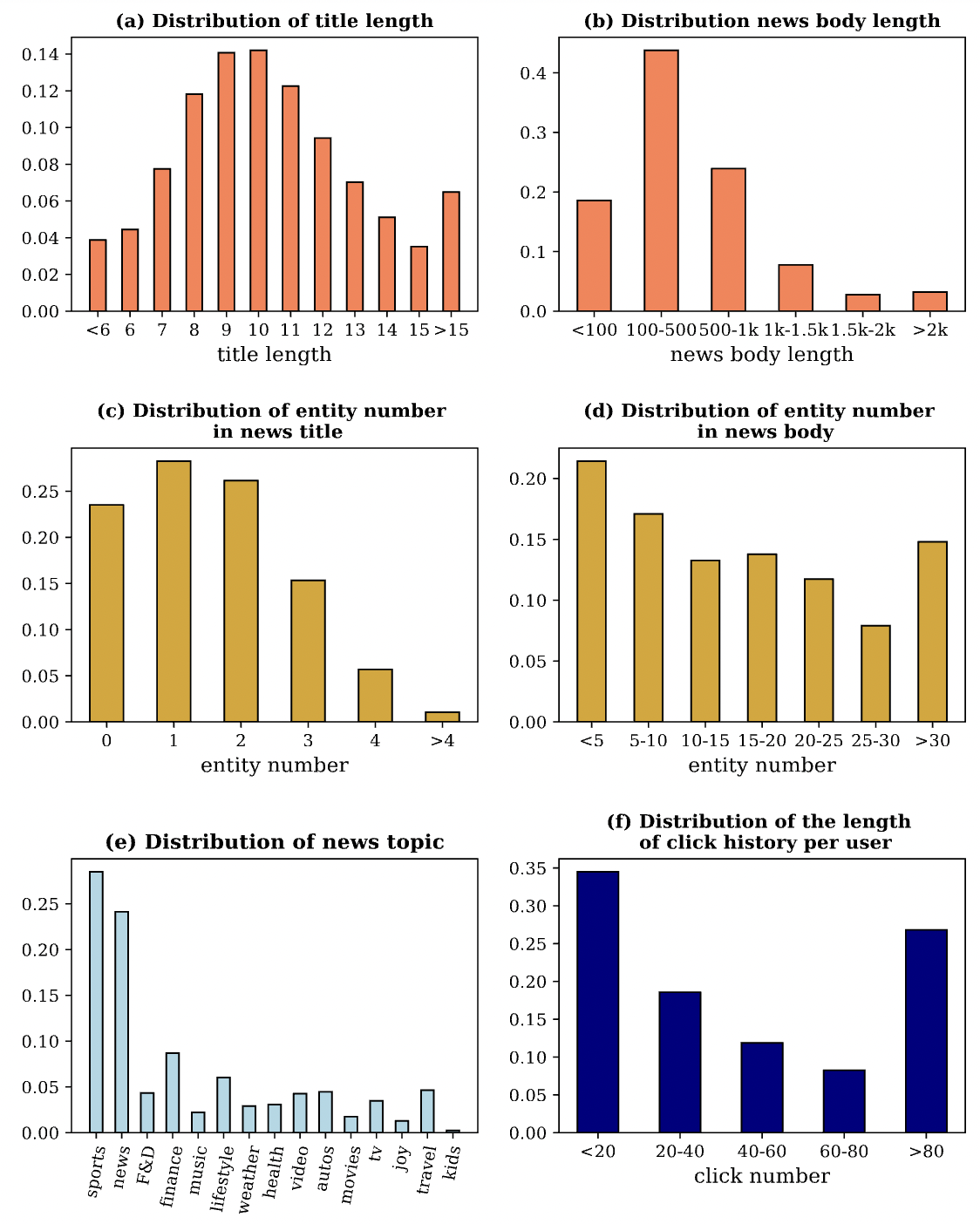}
    \caption{The statistics of news corpus and training set
of the PENS dataset.}
    \label{fig:PENS_Stat}
\end{figure}

\begin{figure*}[t] \
    \centering
    \includegraphics[width = 1\textwidth]{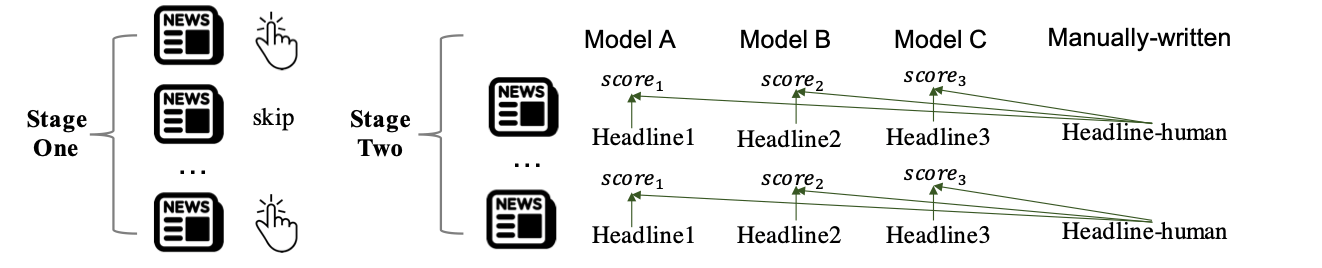}
    \caption{Stages of creation of testing dataset consisting of personalized headlines}
    \label{fig:PENS_Stages}
\end{figure*}

\begin{table*}[t]
\centering
\begin{tabular}{l p{5cm} p{8cm}}
\hline
\textbf{Column} & \textbf{Example Context} & \textbf{Description} \\ \hline
userid & NT1 & The unique ID of 103 users \\ 
clicknewsID & N108480, N38238, N35068, ... & The user’s historical clicked news collected at the first stage \\ 
posnewID & N24110, N62769, N36186, ... & The exhibited news for each user at the second stage \\ 
rewrite\_titles & 'Legal battle looms over Trump EPA's rule change of Obama's Clean Power Plan rule ...' & The manually-written news headlines for the exhibited news articles and can be split by '\#TAB\#' \\ 
\hline
\end{tabular}
\caption{ Dataset Format of data collected from different users consisting of the articles clicked by the user and details of articles for which personalized headlines created by the user}
\label{tab:datasetFormat}
\end{table*}

The PENS dataset is a comprehensive collection of 113,762 news articles, each of which is categorized into one of 15 distinct topics. Each article in the dataset includes a unique news ID, a title, a body, and a category that has been manually tagged by editors. The average length of a news title is 10.5 words, while the average length of a news body is 549.0 words (Refer to Figure \ref{fig:PENS_Stat} for statistics of news articles). Entities from each news title are extracted and subsequently linked to corresponding entities in WikiData.

For the purpose of training, 500,000 user-news impressions were sampled from June 13, 2019, to July 3, 2019. An impression log records the news articles displayed to a user and the user's click behaviors on these articles during a specific visit to the news website. Each labeled sample in the training set follows the format [uID, tmp, clkNews, uclkNews, clkedHis], where 'uID' represents the anonymous ID of a user, 'tmp' denotes the timestamp of the impression record, 'clkNews' and 'uclkNews' are the clicked and un-clicked news in the impression, respectively, and 'clkedHis' represents the news articles previously clicked by the user. All samples in 'clkNews', 'uclkNews', and 'clkedHis' are sorted by the user’s click time.

\subsection{Test Set Construction Process}

In order to establish an offline testbed, 103 English native speakers, all of whom are college students, were invited to manually create a test set in two stages as represented in Figure \ref{fig:PENS_Stages}.

\textbf{In the first stage}, each participant browses 1,000 news headlines and marks at least 50 pieces that they find interesting. These selected news articles were randomly chosen from the news corpus and were arranged according to their first exposure time.

\textbf{In the second stage}, participants are asked to write down their preferred headlines for another 200 unseen news articles from the dataset without being shown the original news titles. They are also asked to highlight important segments in the original news articles. These unseen news articles are evenly sampled, and they are redundantly assigned to ensure that each news article is reviewed by an average of four people. The quality of these manually-written headlines is checked by professional editors from the perspective of the factual aspect of the media frame. Headlines that are of low quality, such as those containing incorrect factual information, those inconsistent with the news body, or those that are too short or too long, are excluded. The remaining headlines are considered to be the personalized reading focuses of the annotators on the articles, and are taken as gold-standard headlines in the PENS dataset.

\begin{figure*}[t]
    \centering
    \includegraphics[width=1\textwidth]{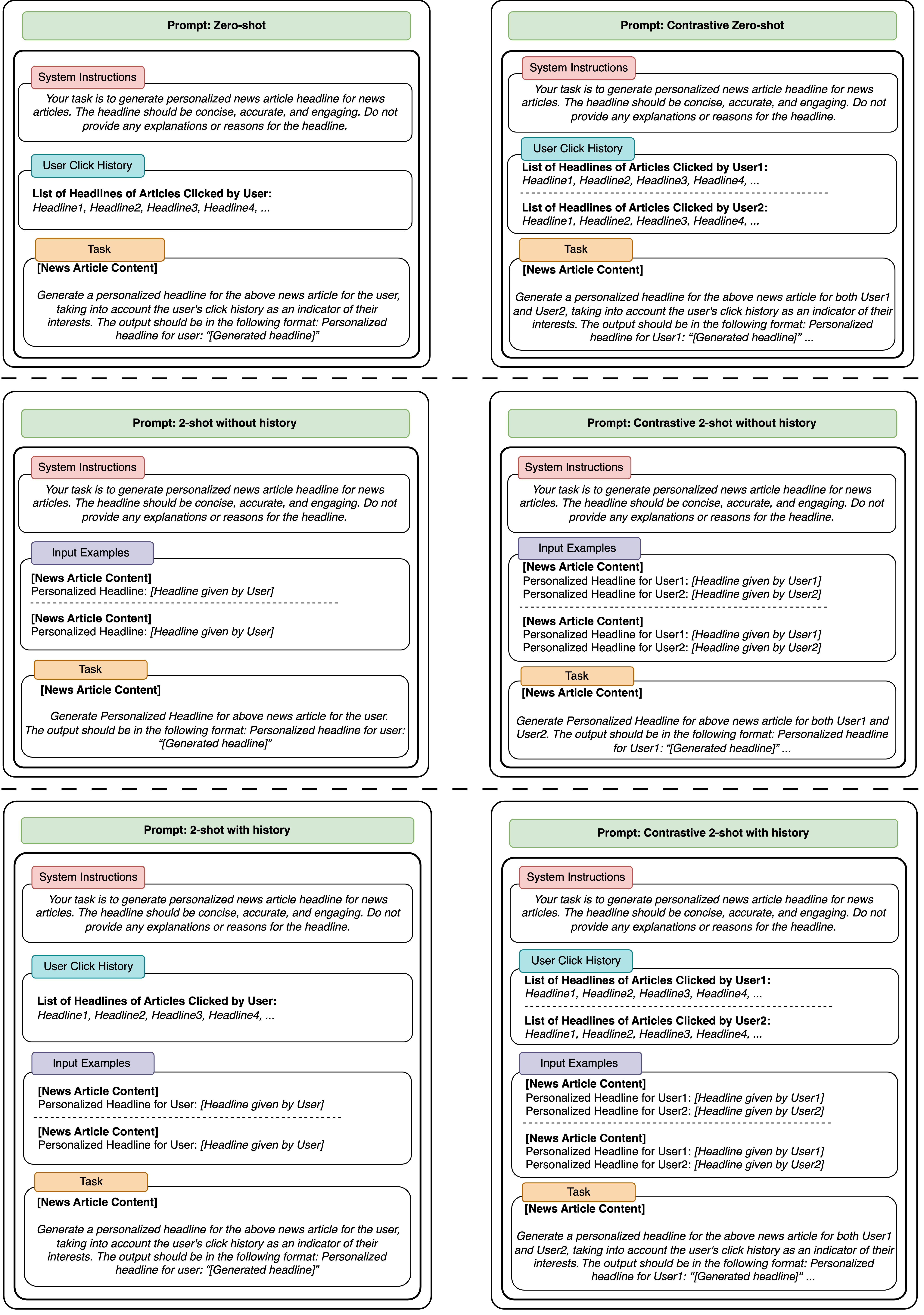}
    \caption{\textbf{Prompt Templates within the \copernicus\ framework}: Prompts on the left probe whether models utilize richer reader profiles; prompts on the right probe whether models utilize contrastive information for real personalization.}
    \label{fig:PromptStructure}
\end{figure*}

\section{Prompt Design Principles}\label{app:prompt design}
We use six different types of prompting styles, each tailored to provide varying levels of context and personalization to the LLMs. We present the structure of each prompt style in Figure \ref{fig:PromptStructure}, along with the composition of tokens in the prompts as described in Table \ref{tab:ContextLengthTable}, to provide a comprehensive understanding of the composition of prompting techniques employed in our study. If the length of any portion of the prompt is greater than the limit, it's truncated to fit it in the context length. For prompt examples refer to Figures \ref{fig:Example PX-1}-\ref{fig:Ex_PX5} in Appendix \footnote{News article statistics from the sample of 3840 news articles used in this study --  The mean length was 659.91 tokens, the median was 493 tokens, the 90th percentile was 1180 tokens, and the 95th percentile was 1659 tokens.}
\begin{enumerate}
    \item \textbf{Zero-shot}: In this approach, we provide the user’s click history followed by the target article for which we aim to generate a personalized headline.
    \item \textbf{Contrastive zero-shot}: This approach provides the click history of two users (User1 and User2), followed by the target article clicked by both users.
    \item \textbf{2-shot w/o history}: In this approach, we provide two example articles and their corresponding headlines given by the user, followed by the target article for which we aim to generate a personalized headline.
    \item \textbf{Contrastive 2-shot w/o history}: This method is similar to the 2-shot approach but involves providing two example articles and their corresponding headlines given by two users (User1 and User2).
    \item \textbf{2-shot w/ history}: In this approach, we provide the user's click history, two examples of articles clicked by the user and their corresponding summaries, along with the article body that needs to be summarized.
    \item \textbf{Contrastive 2-shot w/ history}: This method involves providing the click history for both users, two examples of common articles clicked by both users and their corresponding summaries, along with the article body that needs to be summarized.
\end{enumerate}

In the study, a substantial number of prompts were employed for probing, as indicated in Table \ref{tab:ContextLengthTable}. The generated headlines were extracted from the produced text using simple regular expression matching. The output format was explicitly demonstrated in the examples for 2-shot prompts, and an example format was provided for 0-shot prompts. 

\section{Analysis of Models w.r.t \copernicus} \label{app:evaluation and results}

\subsection{Adversarial Testing: Results}\label{app:adversarial test results}
Adversarial testing was conducted for contrastive prompts across three distinct types of Large Language Models (LLMs): Mistral 7B Inst. v0.2 which exhibits a single paradox; Llama 2 13B Chat that displays a moderate number of paradoxes; and Tulu V2 DPO 7B which presents all paradoxes as shown in Table \ref{tab:paradox table}. The details of User-2, such as their click history and the headlines they wrote in the contrastive prompt, are not accurate. Instead, the reading history of another user and the headline of a random article written by a random user were used. Table \ref{tab:Adversarial validation} verifies the existence of the paradoxes, as these models show higher perplexity for three different types of contrastive prompts: PX-3, which is tested by the contrastive zero-shot prompt; PX-4, which is tested by the contrastive two-shot prompt without history; and PX-5, which is tested by the contrastive two-shot prompt with history. These results indicate that these models are sensitive to the quality and relevance of the information provided in the prompts, and that they perform worse when the prompts contain incorrect details about the user’s reading history or the headlines written by them in the examples. Intriguingly, these three models exhibit a significant performance decline when the prompts include user’s reading history. However, they only show a minor performance drop for contrastive two-shot prompts without history. This suggests that these models are capable of discerning that an incorrect/irrelevant headline is given by User-2 in the examples.

\begin{table}[h]
    \centering
    \scalebox{0.66}{
    \begin{tabular}{lccc}
        \toprule
        \hline
        \textbf{Models}&\textbf{C-0-shot}&\textbf{C-2-shot w/o hist} &\textbf{C-2-shot w/ hist}\\
        \midrule
        \hline
        Llama 2 13B Chat& $q_1$ & $q_2$ & $q_3$\\
        
        Mistral 7B Inst. v0.2 & $q_4$ & $q_5$ & $q_6$\\
        
        Tulu V2 DPO 7B & $q_7$ & $q_8$ & $q_9$\\
        \hline
    \end{tabular}}
    \caption{\textbf{(Survey) Questionnaire Structure:} A respondent fills up the survey for document $d_{i}$ in the sequence: $\langle u_{ij}, u_{ik}\rangle \rightarrow q_{i1} \rightarrow q_{i2} \rightarrow \ldots \rightarrow q_{i9}$, where $q_{i\bullet}: \langle s_{u_{ij}}, s_{u_{ik}} \rangle^{(M_{\boldsymbol{\theta},u},\mathcal{P_C})_\bullet}$; $s_{u}$ is the summary generated by each of the 3 models for a specific prompt type (i.e., the model-contrastive prompt-type pair $(M_{\boldsymbol{\theta},u},\mathcal{P_C})_\bullet$).}  
    \label{tab:HJ Survey Questions}
    \vspace{-5mm}
\end{table}

\begin{table}[t]
    \centering
    \scalebox{0.8}{
    \begin{tabular}{lcc}
        \toprule
        \hline
        \textbf{Models}&\textbf{PX-5}&\textbf{PX-6}\\
        \midrule
        \hline
        Llama 2 13B Chat& \textcolor{teal}{\cmark}& \textcolor{teal}{\cmark}\\
        
        Mistral 7B  Inst. v0.2&\textcolor{red}{\xmark}&\textcolor{teal}{\cmark}\\
        
        Tulu V2 DPO 7B &\textcolor{teal}{\cmark}&\textcolor{teal}{\cmark}\\
        \hline
    \end{tabular}
    }
    \caption{\textbf{Human-Judgment Validation:} PX-5 exists for worst and medium case; Contrastive PX-1 (PX-6) is also confirmed (Human-agreement denoted as \textcolor{teal}{\cmark}).}
    \label{tab:HJ Validation}
    \vspace{-5mm}
\end{table}

\subsection{Human Judgment Validation: Results}\label{app:HJ Validation}
\paragraph{Survey Structure:} Human judgment-based validation was conducted on a set of contrastive prompts, systematically evaluating three distinct LLMs: (i) Mistral 7B Instruct v0.2 (exhibiting only PX-5), (ii) Llama 2 13B Chat (exhibiting a moderate number of paradoxes - PX-1/2/5), and (iii) Tulu V2 DPO 7B (exhibiting all the paradoxes), as outlined in Table \ref{tab:paradox table}. We have randomly sampled multiple documents and three corresponding users (i.e., readers) who have generated summaries for those documents eg.$(d_{i},(u_{i1},u_{i2},u_{i3}))$ from the PENS dataset. For each $(d_{i},(u_{i1},u_{i2},u_{i3}))$ there are 3 combinations possible $((d_{i},(u_{i1},u_{i2}))$, $(d_{i},(u_{i2},u_{i3}))$, $(d_{i},(u_{i3},u_{i1})))$. Each survey respondant is shown a set of 10 questions $(\langle u_{ij}, u_{ik}\rangle,q_{i1},q_{i2}...q_{i9})$ for a given combination corresponding to a document $d_{i}$ eg.$(d_{i},(u_{ij},u_{ik}))$ as shown in the Table \ref{tab:HJ Survey Questions}; where $q_{i\bullet} \in \left \{ q_{i1},q_{i2}...q_{i9} \right \}$  contains a pair of the corresponding prompts and model generated summaries ($\langle s_{u_{ij}}, s_{u_{ik}} \rangle^{(M_{\boldsymbol{\theta},u},\mathcal{P_C})_\bullet}$) of user-pairs and $\langle u_{ij}, u_{ik}\rangle$ contains a pair of user-generated reference summaries of user-pairs.\\
\vspace{-4mm}
\paragraph{Survey details:} We designed and conducted an online survey to gather insights from participants on a voluntary basis (see Figure \ref{fig:Ex_Survey} in Appendix). A total of 339 responses were obtained, encompassing evaluations of 113 documents from the PENS evaluation dataset. The respondent pool comprised 262 males and 77 females. Participants represented diverse educational backgrounds, including undergraduate and graduate students specializing in computer science, mathematical science, electronic engineering, and humanities. To maintain objectivity, participants were not informed that the questions presented were summary-pairs. Instead, each summary pair was displayed as regular text, prompting participants to rate their similarity on a scale ranging from 1 (low) to 6 (very high). This methodology ensured unbiased judgments regarding the proximity of subjective user reference summaries and their corresponding model-generated summaries. 

\paragraph{Computing EGISES-HJ-JSD:} The similarity scores provided by users were utilized as the basis for computing similarity scores for summary-summary pairs. For summary-document distance, we used JSD (since it is not viable for a user to read the whole document and assign scores to a summary). Using these distances (summary-summary distances based on user ratings, summary-document distances using JSD), we then calculated EGISES-HJ-JSD for all three models and prompt types. We observed that EGISES-HJ-JSD scores based on human judgment showed agreement with \textbf{2/3} models for PX-5 and \textbf{3/3} models for PX-6 (the contrastive version of PX-1) as shown in Table \ref{tab:HJ Validation}.

\section{Inference Setup and Configuration}\label{app:inference setup}
For the inference process in our study, we utilized the JAX library. Our implementation heavily relied on the repository "llama-2-jax" \citep{llama2jaxAyaka}. In order to generate outputs from our models, we configured the inference process to ensure optimal performance and output quality.

\subsection{LLM Settings} We used the following settings to control the behavior of the LLMs during inference:

\begin{itemize}
\itemsep0em
\item \textbf{Temperature:} We set the temperature to a value of 0.6, striking a balance between predictability and diversity.
\item \textbf{Sampling Method:} We adopted the top-k sampling method with k = 16 for generating outputs.
\item \textbf{Maximum Length:} Maximum length for inference was set to the 4096 tokens.
\item \textbf{Inference Precision:} We used bfloat16 precision for inference, consistent with the precision in which the model weights were originally published.

\end{itemize}

\subsection{Compute Platform} For high-performance inference, we utilized TPU v3-8/v4-8 VMs through Google Cloud Platform. 

\begin{figure*}
    \centering
    \includegraphics[width=1\textwidth]{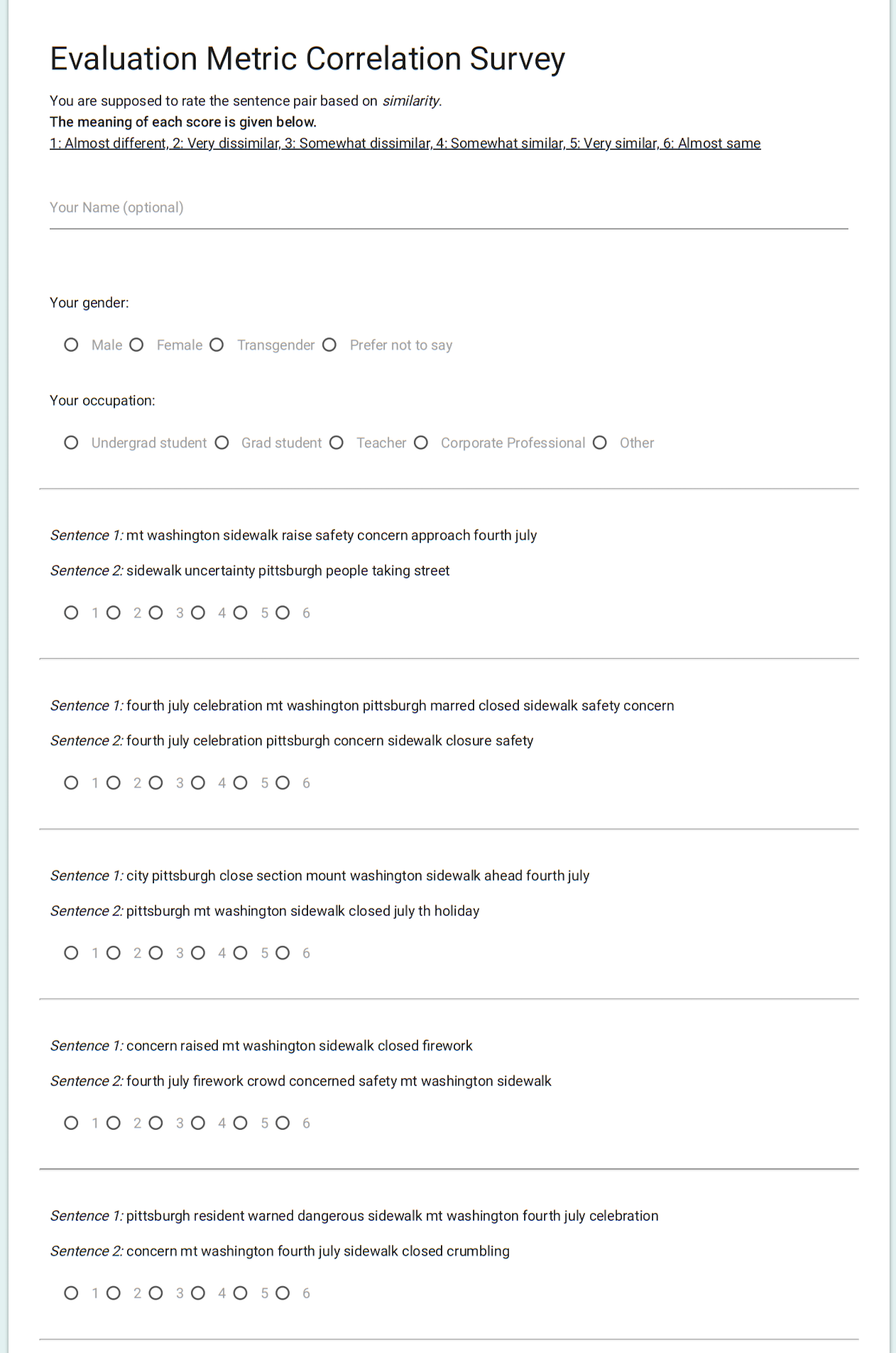}
    \caption{Website portal designed for conducting survey for collecting human judgements on the similarity between user references and model generated summaries for contrastive prompts.}
    \label{fig:Ex_Survey}
\end{figure*}

\begin{figure*}
    \centering
    \includegraphics[width=0.88\textwidth]{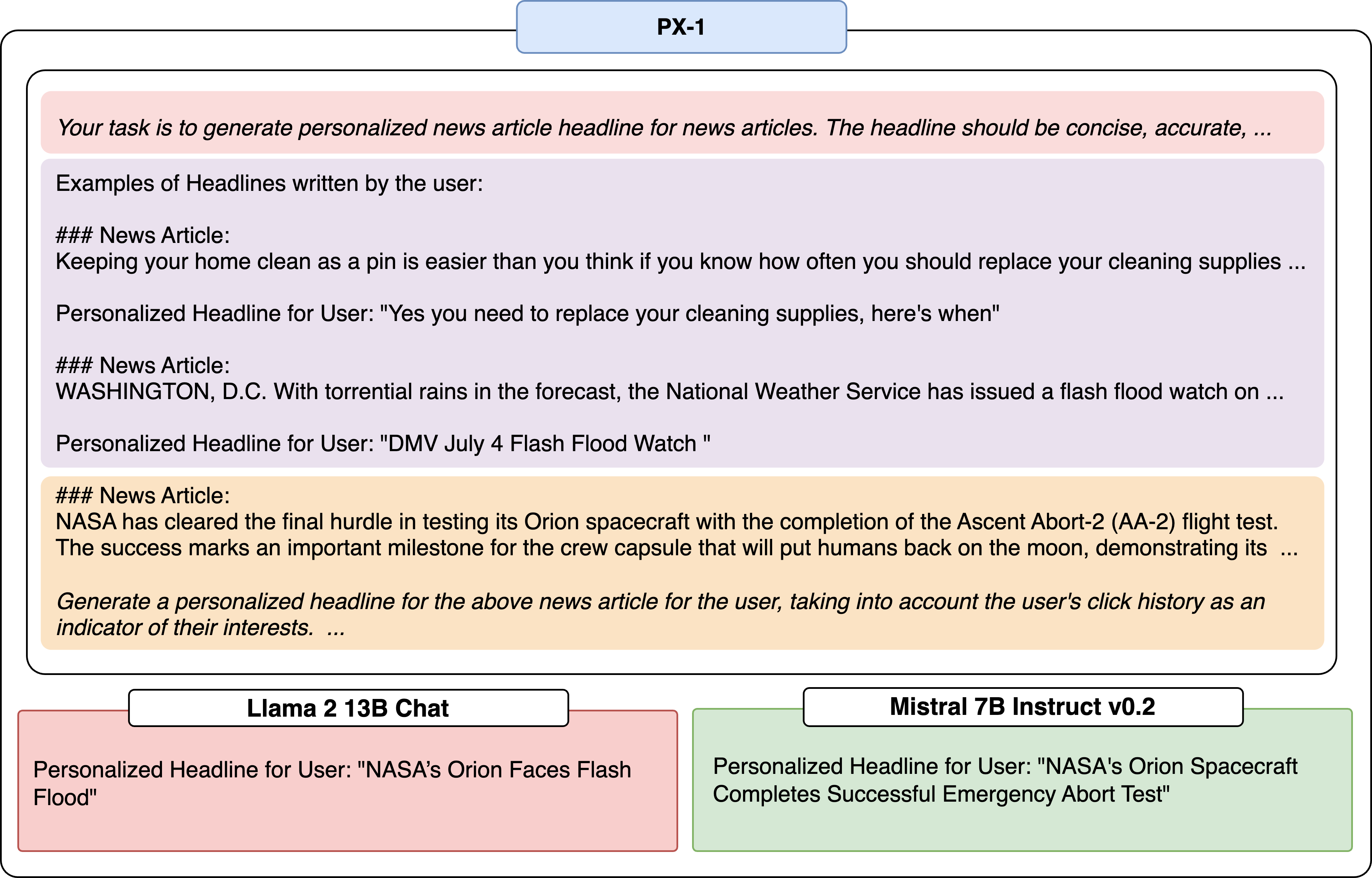}
    \caption{\textbf{Illustration of PX-1} (\textit{effect of personalized examples}):  The left column shows the output of Llama 2 13B Chat, which hallucinates generating irrelevant information (marked in red) due to the distraction caused by the examples; the right column shows the output of Mistral 7B Instruct v0.2, which generates the expected response (marked in green).}
    \label{fig:Example PX-1}
\end{figure*}

\begin{figure*}
    \centering
    \includegraphics[width=0.88\textwidth]{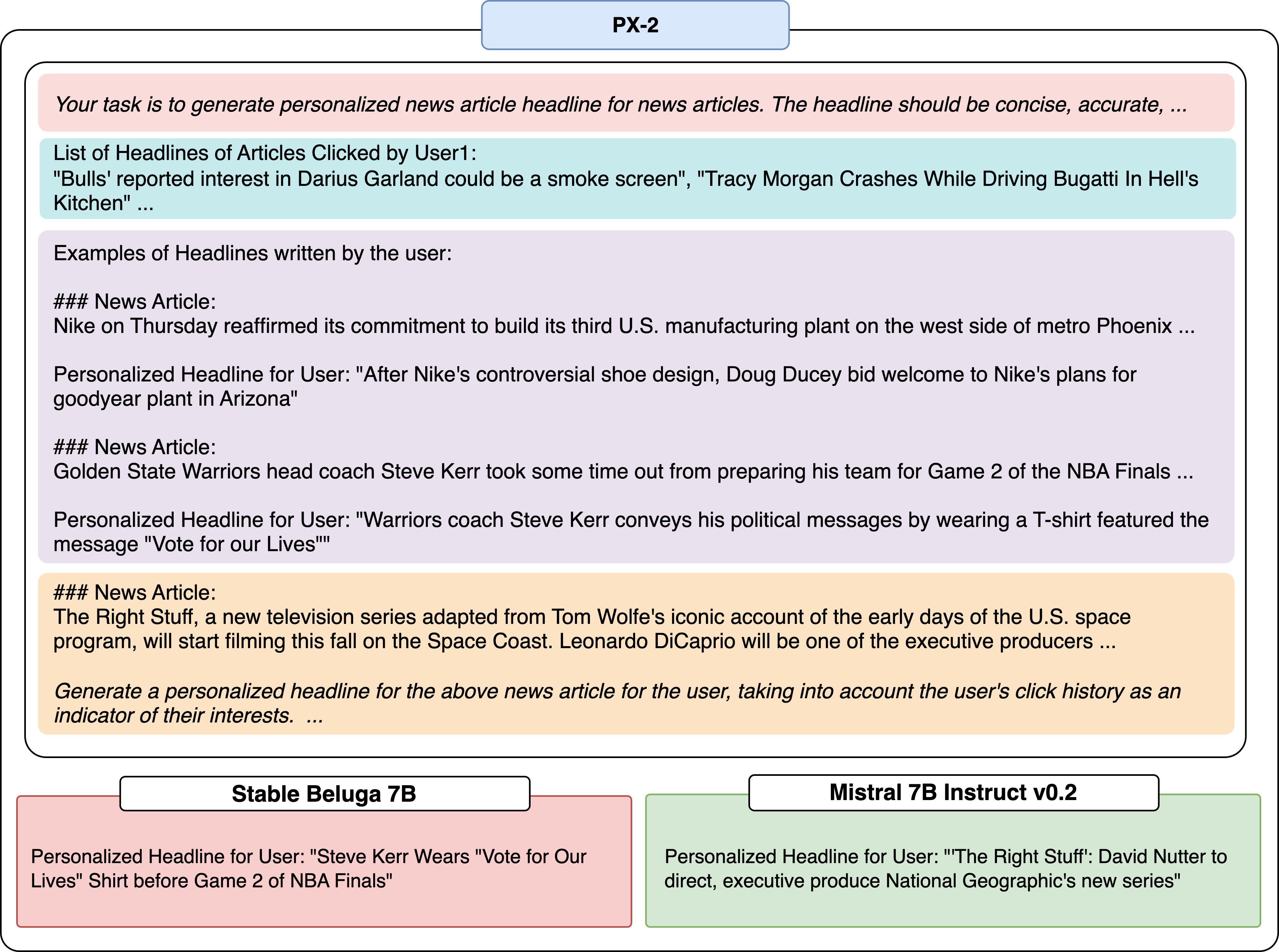}
    \caption{\textbf{Illustration of PX-2} (\textit{effect of personalized headline click history}): The left column shows the output of Stable Beluga 7B, which hallucinates incorrect information (marked in red) due to the inability to reinforce historical interest on TV-series with the current interest (i.e., query concepts); the right column shows the output of Mistral 7B Instruct v0.2, which generates the expected response (marked in green).}
    \label{fig:Ex_PX2}
\end{figure*}

\begin{figure*}
    \centering
    \includegraphics[width=0.9\textwidth]{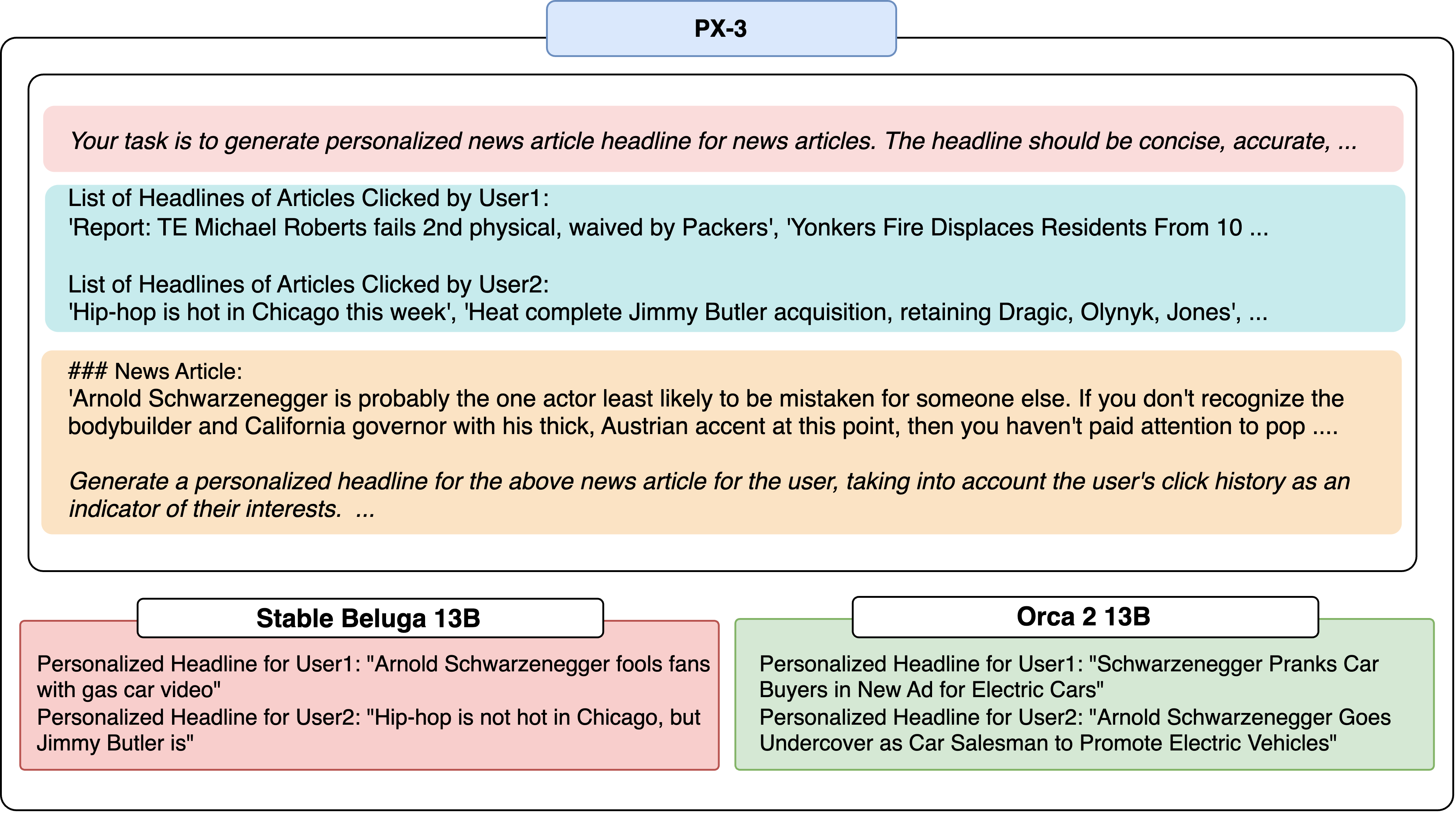}
    \caption{\textbf{Illustration of PX-3} (\textit{effect of \underline{contrastive} personalized click history}): The left column shows the output of Stable Beluga 13B, which hallucinates inaccurate information (marked in red) due to the distraction caused by the list of articles clicked by two users. The right column shows the output of Orca 2 13B, which generates the expected response (marked in green).}
    \label{fig:Ex_PX3}
\end{figure*}

\begin{figure*}
    \centering
    \includegraphics[width=0.9\textwidth]{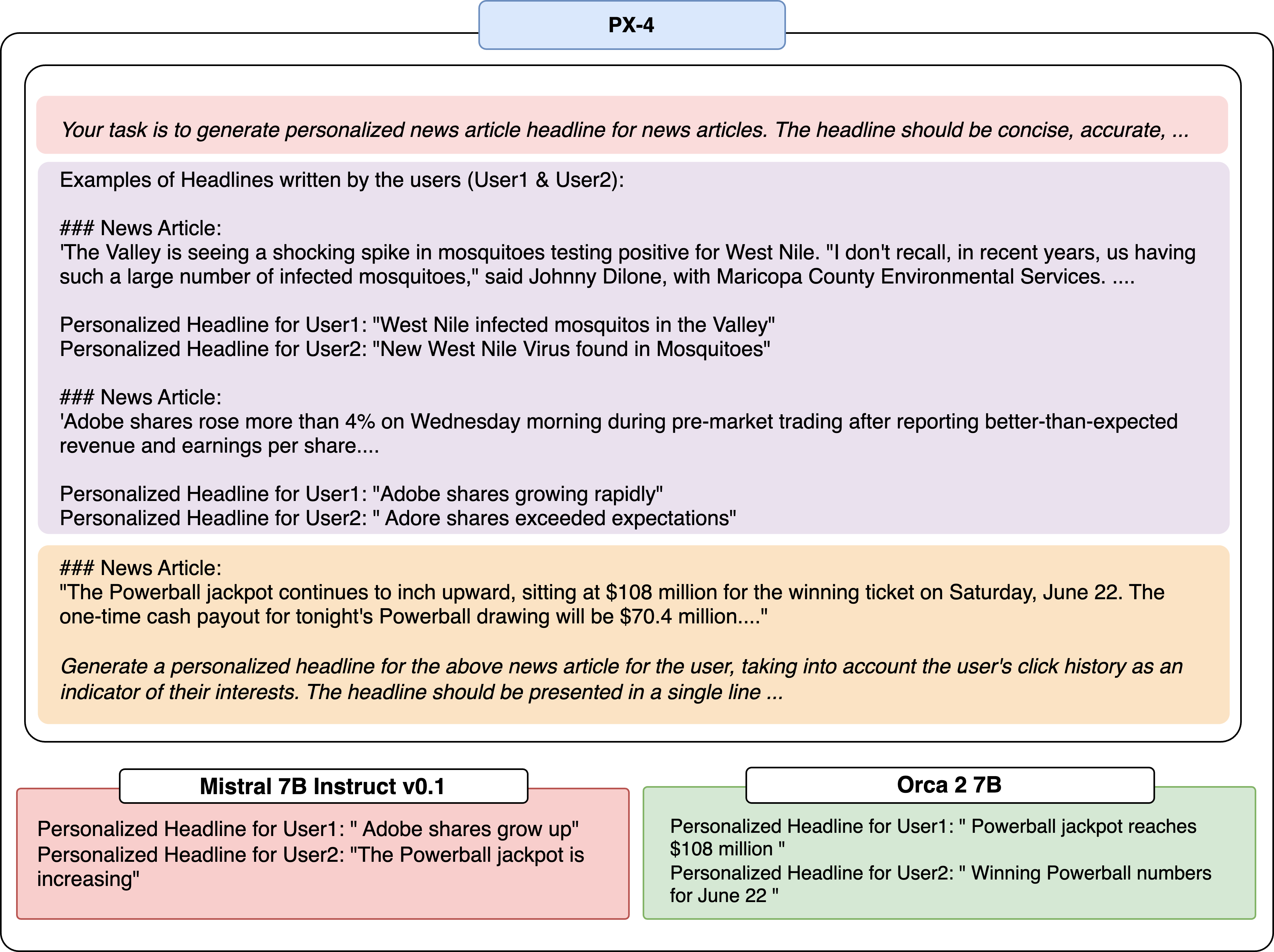}
    \caption{\textbf{Illustration of PX-4} (\textit{effect of \underline{contrastive} personalized examples}): The left column shows the output of Mistral 7B Instruct v0.1, which hallucinates inconsistent information (marked in red) due to the distraction resulting from the unintended reinforcement of common concepts (\textit{adobe}, \textit{share}, \textit{grow}) in both the contrastive personalized headline examples provided by different users that overflows to the response of the query document; the right column shows the output of Orca 2 7B, which generates the expected response (marked in green) and does not suffer from any such overflow.} 
    \label{fig:Ex_PX4}
\end{figure*}

\begin{figure*}
    \centering
    \includegraphics[width=0.9\textwidth]{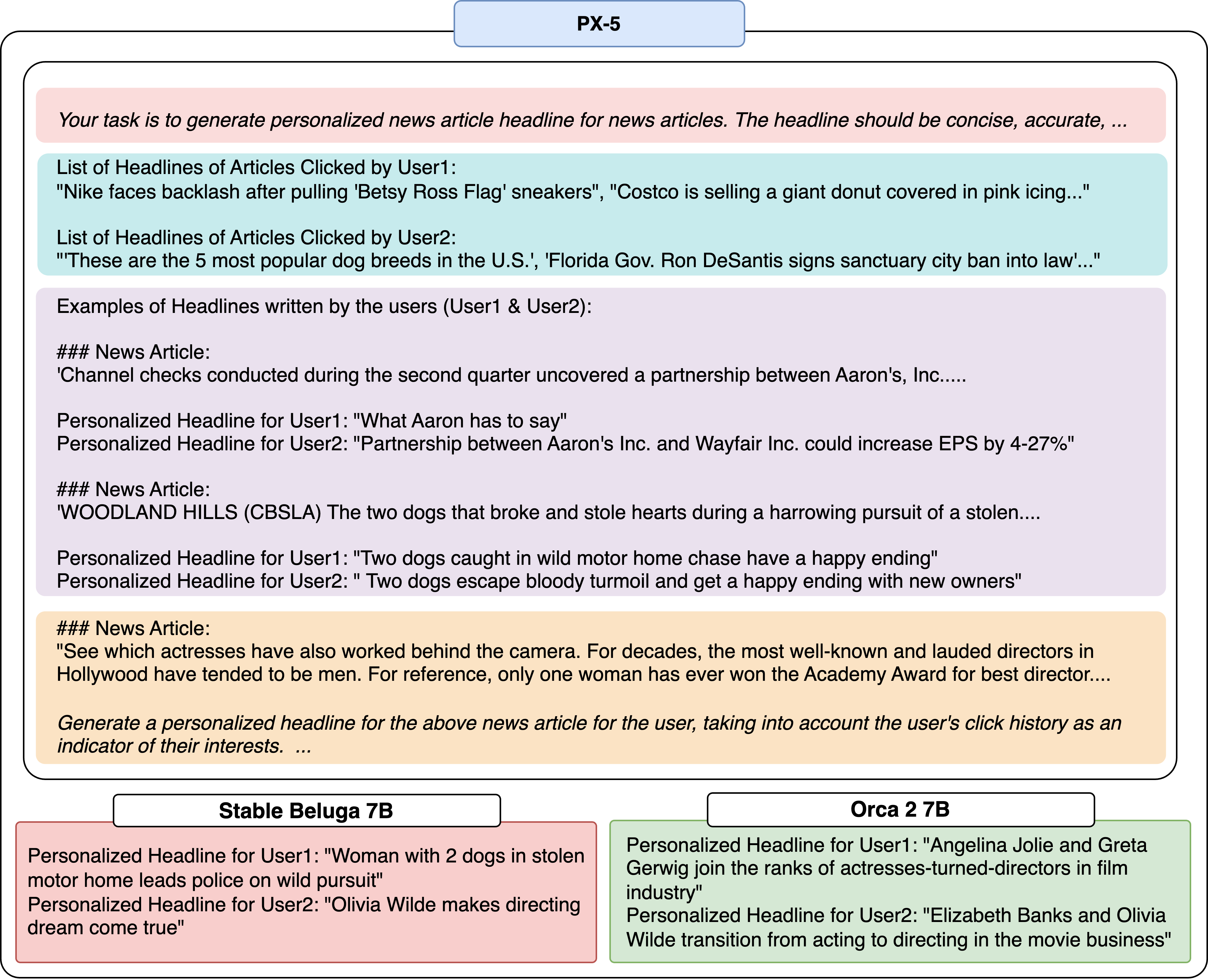}
    \caption{\textbf{Illustration of PX-5} (\textit{effect of \underline{contrastive} personalized examples \underline{with click history}}): The left column shows the output of Stable Beluga 7B, which hallucinates irrelevant information (marked in red) due to the distraction caused by the cross-association of concepts in the click history of user-1 with that of user-2; the right column shows the output of Orca 2 7B, which generates the expected response (marked in green).}
    \label{fig:Ex_PX5}
\end{figure*}

\end{document}